\documentclass[letterpaper,journal]{IEEEtran}
\usepackage{amsmath,amsfonts}
\usepackage{algorithmic}
\usepackage{algorithm}
\usepackage{array}
\usepackage[caption=false,font=normalsize,labelfont=sf,textfont=sf]{subfig}
\usepackage{textcomp}
\usepackage{stfloats}
\usepackage{url}
\usepackage{verbatim}
\usepackage{graphicx}
\usepackage{cite}
\usepackage{mathtools}
\usepackage{amsmath}
\usepackage{amssymb}
\usepackage{bm}
\usepackage[dvipsnames]{xcolor}
\usepackage{glossaries}
\usepackage{makecell}
\usepackage{siunitx}
\usepackage{enumitem}

\DeclareMathOperator*{\argmin}{argmin}

\newcommand{\pare}[1]{\left({#1}\right)}

\newcommand{\abs}[1]{\lvert#1\rvert}

\newcommand{\pitwo}{\frac{\pi}{2}}

\DeclareMathOperator{\FK}{FK}
\DeclareMathOperator{\ID}{ID}
\DeclareMathOperator{\Rot}{R}

\newcommand{\s}{s}
\newcommand{\ds}{ds}

\newcommand{\dt}{dt}

\newcommand{\q}{\bm{q}}
\newcommand{\dq}{\dot{\q}}
\newcommand{\ddq}{\ddot{\q}}
\newcommand{\torque}{\bm{\tau}}

\newcommand{\dqi}{\dot{q}_i}
\newcommand{\ddqi}{\ddot{q}_i}
\newcommand{\torquei}{\tau_i}

\newcommand{\dqLimit}{\bar{\dq}}
\newcommand{\ddqLimit}{\bar{\ddq}}
\newcommand{\torqueLimit}{\bar{\torque}}

\newcommand{\dqLimiti}{\bar{\dot{q}}_i}
\newcommand{\ddqLimiti}{\bar{\ddot{q}}_i}
\newcommand{\torqueLimiti}{\bar{\tau}_i}

\newcommand{\qt}{\q(t)}
\newcommand{\dqt}{\dq(t)}
\newcommand{\ddqt}{\ddq(t)}

\newcommand{\dsdt}{\frac{\ds}{\dt}}

\newcommand{\qz}{\q_0}
\newcommand{\dqz}{\dot{\q}_0}
\newcommand{\ddqz}{\ddot{\q}_0}

\newcommand{\qd}{\q_d}
\newcommand{\dqd}{\dot{\q}_d}

\newcommand{\Path}{\bm{p}}
\newcommand{\Paths}{\bm{p}(s)}
\newcommand{\dPaths}{\frac{d\Paths}{\ds}}
\newcommand{\ddPaths}{\frac{d^2\Paths}{\ds^2}}

\newcommand{\tfroms}{\psi(s)}

\newcommand{\overdt}{\frac{d}{\dt}}

\newcommand{\rate}{r}
\newcommand{\rates}{\rate(s)}
\newcommand{\drates}{\frac{d\rates}{\ds}}

\newcommand{\traj}{\bm{\zeta}}

\newcommand{\Rplus}{\mathbb{R}_+}
\newcommand{\manifold}{\mathcal{M}}
\newcommand{\Loss}{\mathcal{L}}
\newcommand{\nnLoss}{L}
\newcommand{\stepNNLoss}{L_t}

\newcommand{\taskSpaceManifold}{\mathcal{T}}

\newcommand{\vtheta}{\bm{\theta}}

\newcommand{\knots}{\bm{k}}

\newcommand{\qControlPoints}{\mathcal{C}^{\Path}}
\newcommand{\tControlPoints}{\mathcal{C}^{\rate}}

\newcommand{\numControlPoints}{C}
\newcommand{\bsplineDegree}{D}

\newcommand{\numQControlPoints}{\numControlPoints_{\Path}}
\newcommand{\numTControlPoints}{\numControlPoints_{\rate}}

\newcommand{\qDegree}{\bsplineDegree_{\Path}}
\newcommand{\tDegree}{\bsplineDegree_{\rate}}

\newcommand{\NNPathOutput}{\phi^{\Path}}

\newcommand{\Lm}{\bm{\lambda}}
\newcommand{\lm}{\lambda}
\newcommand{\constraint}{c}
\newcommand{\Constraint}{\bm{c}}
\newcommand{\slack}{\mu}
\newcommand{\Slack}{\bm{\mu}}
\newcommand{\bLM}{\bar{C}}
\newcommand{\bLMi}{\bar{C}_i}
\newcommand{\cLMi}{{\Loss}_{\manifold,\Lambda}^{\neg i}}
\newcommand{\stepLoss}{\Loss_{t}}
\newcommand{\manifoldLoss}{\Loss_\manifold}

\newcommand{\manifoldMetricLoss}{\Loss_{\manifold, \Lambda}}
\newcommand{\stepManifoldLoss}{\Loss_{\manifold_t}}
\newcommand{\stepManifoldLossi}{\Loss_{\manifold_t}^i}
\newcommand{\alphas}{\bm{\alpha}}
\newcommand{\balpha}{\bar{\alpha}}

\newcommand{\diag}{\operatorname{diag}}

\DeclareMathOperator{\huber}{H}
\DeclareMathOperator{\relu}{ReLU}
\DeclareMathOperator{\ind}{I}

\newcommand{\ndof}{n_{dof}}
\newacronym{rl}{RL}{Reinforcement Learning}
\newacronym{drl}{DRL}{Deep Reinforcement Learning}
\newacronym[plural=MDPs, firstplural=Markov Decision Processes (MDPs)]{mdp}{MDP}{Markov Decision Process}
\newacronym{kl}{KL}{Kullback-Leibler Divergence}

\newacronym{vae}{VAE}{Variational Auto-Encoders}

\newacronym{mom}{MOM}{Measure of Manipulability}
\newacronym{nl}{NLP}{Nonlinear Programming}
\newacronym{lp}{LP}{Linear Programming}
\newacronym{qp}{QP}{Quadratic Programming}
\newacronym{aqp}{AQP}{Anchored Quadratic Programming}
\newacronym{slsqp}{SLSQP}{Sequential Least Squares Programming}

\newacronym{mpc}{MPC}{Model Predictive Control}

\newacronym{sdf}{SDF}{Signed Distance Function}
\newacronym{tsdf}{TSDF}{Truncated Signed Distance Function}
\newacronym{esdf}{ESDF}{Euclidean Signed Distance Function}
\newacronym{redsdf}{ReDSDF}{Regularized Deep Signed Distance Fields}
\newacronym{apf}{APF}{Artificial Potential Fields}
\newacronym{rmp}{RMP}{Riemannian Motion Policies}
\newacronym{hri}{HRI}{Human-Robot Interaction}
\newacronym{ecomann}{ECoMaNN}{Equality Constraint Manifold Neural Network}
\newacronym{smpl}{SMPL}{A Skinned Multi-Person Linear Model}
\newacronym{poi}{PoI}{Points of Interest}
\newacronym{wbc}{WBC}{whole-body control}

\newacronym{cspace}{C-space}{Configuration Space}

\newacronym{rnea}{RNEA}{ recursive Newton-Euler algorithm}

\newacronym{rrt}{RRT}{Rapidly Exploring Random Trees}
\newacronym{cbirrt}{CBiRRT}{Constrained Bi-directional RRT}

\newacronym{lqr}{LQR}{Linear Quadratic Regulator}
\newacronym{cnn}{CNN}{Convolutional Neural Network}

\newacronym{ours}{CNP-B}{Constrained Neural motion Planning with B-splines}

\newacronym{mpcmpnet}{MPC-MPNet}{Model-Predictive Motion Planning Network}
\newacronym{cem}{CEM}{Cross Entropy Method}

\newacronym{sst}{SST}{Stable Sparse RRT}

\begin{document}

\title{Fast Kinodynamic Planning on the Constraint Manifold with Deep Neural Networks}


\author{Piotr Kicki$^{1}$, Puze Liu$^{2}$, Davide Tateo$^{2}$, Haitham Bou-Ammar$^{3}$,\\ Krzysztof Walas$^{1}$, Piotr Skrzypczy\'nski$^{1}$ and Jan Peters$^{2,4,5}$ 
\thanks{$^{1}$Institute of Robotics and Machine Intelligence, Poznan University of Technology, Poland  {\tt\small \{piotr.kicki, krzysztof.walas, piotr.skrzypczynski\}@put.poznan.pl}}%
\thanks{$^{2}$Department of Computer Science, Technische Universit\"at Darmstadt, Germany {\tt\small puze@robot-learning.de, \{davide.tateo, jan.peters\}@tu-darmstadt.de}}%
\thanks{$^{3}$Huawei R\&D London, United Kingdom \newline {\tt\small haitham.ammar@huawei.com} }%
\thanks{$4$ German Research Center for AI (DFKI), Research Department: Systems AI for Robot Learning}
\thanks{$5$ Hessian.AI}
}

\markboth{Transactions on Robotics,~Vol.~X, No.~Y, Month~Year}%
{Kicki \MakeLowercase{\textit{et al.}}: Fast Kinodynamic Planning on the Constraint Manifold with Deep Neural Networks}

\IEEEpubid{0000--0000/00\$00.00~\copyright~2022 IEEE}

\maketitle

\begin{abstract}

Motion planning is a mature area of research in robotics with many well-established methods based on optimization or sampling the state space, suitable for solving kinematic motion planning.
However, when dynamic motions under constraints are needed and computation time is limited, fast kinodynamic planning on the constraint manifold is indispensable.
In recent years, learning-based solutions have become alternatives to classical approaches, but they still lack comprehensive handling of complex constraints, such as planning on a lower-dimensional manifold of the task space while considering the robot's dynamics.
This paper introduces a novel learning-to-plan framework that exploits the concept of constraint manifold, including dynamics, and neural planning methods.
Our approach generates plans satisfying an arbitrary set of constraints and computes them in a short constant time, namely the inference time of a neural network. This allows the robot to plan and replan reactively, making our approach suitable for dynamic environments.
We validate our approach on two simulated tasks and in a demanding real-world scenario, where we use a Kuka LBR Iiwa 14 robotic arm to perform the hitting movement in robotic Air Hockey.

\end{abstract}


\section{Introduction}

\IEEEPARstart{R}{obotic} manipulators tackle a wide variety of complex tasks in dynamic environments such as ball-in-a-cup~\cite{kawato1994teaching,kober2008policy}, table tennis~\cite{mulling2011biomimetic,buchler2022learning}, juggling~\cite{ploeger2021high}, diabolo~\cite{von2021analytical}, and air hockey~\cite{namiki2013hierarchical, liu2021efficient}. These tasks require quick computation of a feasible trajectory, i.e. a trajectory solving the task while respecting all kinematic and dynamic constraints such as joint position, velocity, acceleration, and torque limits.
On top of that, many tasks define a set of safety or task-specific constraints that must be enforced on the planned solutions.

Although there have been attempts to solve these problems using optimization and sampling-based motion planning algorithms, they all have significant limitations, such as long planning time, computing hard-to-follow plans, or inability to satisfy boundary conditions. 
One of the approaches to the planning on the constraint manifold is to represent it as a collision-free space, which makes sampling valid states highly improbable~\cite{kingston2018sampling, kingston2019exploring}, resulting in considerably increased planning time.
To address this issue, Berenson et al.~\cite{berenson2009manipulation} introduced the concept of projecting the states onto the constraint manifold. However, this approach is restricted to plan paths, which can be hard to follow, as they disregard the system's dynamics.
Recent work of Bordalba et al.~\cite{Bordalba2021atlasKinodynamicRRT} proposes to solve constrained kinodynamic motion planning problems by building a topological atlas of the constraint manifold and then using \gls{lqr} to control the system locally.
Even though this approach allows for solving complex problems, the motion planning times are prohibitively long for dynamic tasks like the game of air hockey or table tennis. Lengthy planning is also noticeable for optimization-based motion planners, which slow down significantly when subjected to highly non-linear constraints~\cite{Xie2020kinodymanicFactorGraphs} and are prone to get stuck in local minima.
An efficient alternative to the above-mentioned methods can be to build a reduced actuator space that keeps the system on states satisfying the constraints~\cite{liu2021robot}.
However, this approach may have problems with the satisfaction of boundary constraints.

\begin{figure}[t!]
    \centering
    \includegraphics[width=\linewidth]{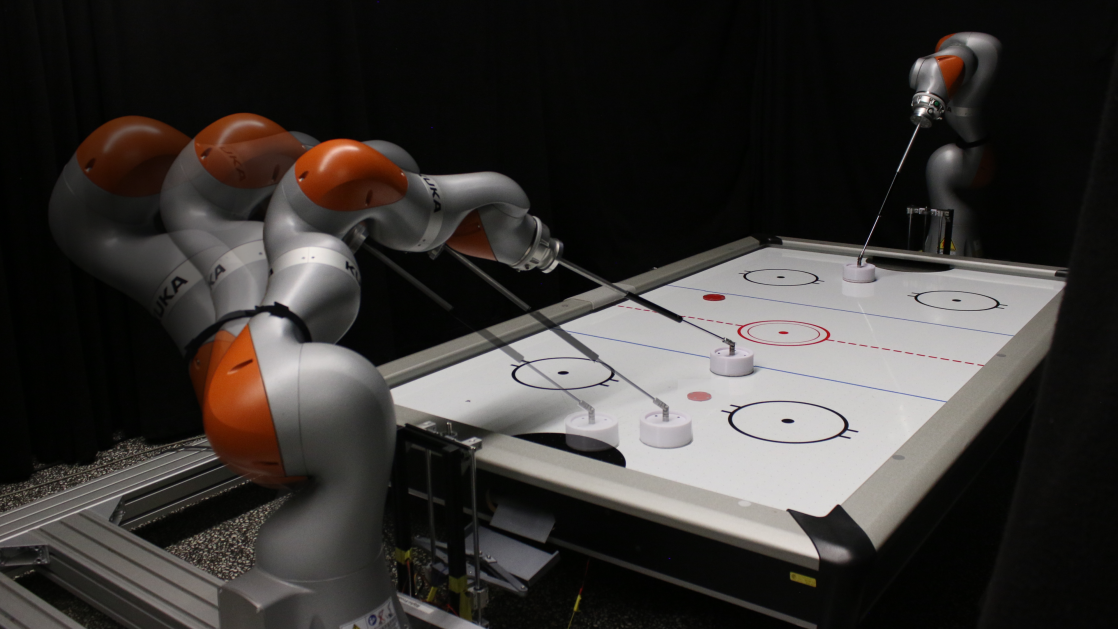}
    \caption{Proposed learning-based motion planning approach enables rapid planning and replanning of smooth trajectories under complex kinodynamic constraints in dynamic scenarios, e.g. the robotic Air Hockey hitting.}
    \label{fig:cover}
\end{figure}

\IEEEpubidadjcol

We propose a novel learning-based approach for constrained kinodynamic planning that efficiently solves all the above-mentioned problems, called \gls{ours}. Similarly to~\cite{berenson2009manipulation}, we frame the problem as planning on a constraint manifold $\mathcal{M}$. Using a similar derivation to~\cite{liu2021robot}, we define all the kinematics, dynamics, and safety/task constraints as a single constraint manifold, defined as the zero level set of an arbitrary constraint function $c(\q,\dq,\ddq)$.
Our approach does not use an online projection~\cite{berenson2009manipulation} nor continuation~\cite{Bordalba2021atlasKinodynamicRRT}, instead, it exploits the representation power of Deep Neural Networks to learn a planning function~\cite{kicki2022bspline}, indirectly encoding the manifold structure. Implicitly learning the manifold enables us to rapidly plan smooth, dynamically feasible trajectories under arbitrary constraints and highly dynamic motions (see Fig.~\ref{fig:cover}). Differently from~\cite{liu2021robot}, our approach does not require building an abstract action space, making it easy to exploit standard planning heuristics and improving the interpretability of the plan. In our work, we frame constraint satisfaction as a manifold learning problem, where the planning function is encouraged to generate trajectories that minimize not only some arbitrary task cost but also the distance from the constraint manifold. 
Furthermore, violating particular constraints to a certain degree can have only a minor negative impact, while violating other constraints can be unacceptable or dangerous.  
Hence, we recover the metric of the constraints manifold, which allows us to attribute different priorities to different constraints.

\subsection{Contributions}
We propose a novel learning-to-plan method for fast constrained kinodynamic planning, which is based on a deep neural network generating trajectories represented using two B-spline curves. 
This paper presents the following contributions:
\begin{enumerate}[label=\roman*., leftmargin=10pt]
\item A new approach to robotic arm motion planning that enables planning dynamic trajectories under constraints. 
This approach is inspired by two ideas from our recent works on car maneuver planning: learning from experience with a deep neural network~\cite{kicki2021learning} and using B-splines for efficient representation of the learned path~\cite{kicki2022bspline}. 
This work applies these concepts to planning a trajectory on a lower-dimensional manifold of the task space while considering the robot's dynamics.
To this end, we extend the learning system architecture to learn Lagrangian multipliers in the optimization problem, learning simultaneously a metric of our constraint manifold.
This novel approach allows us to weigh each constraint by how much it is important to find a feasible solution to the planning problem. 
We demonstrate in simulations and experiments that this new approach is not only much faster than all state-of-the-art methods we were able to compare as baselines, but generates also trajectories that allow faster and more accurate robot motion while being executed\footnote{Code and data can be found at \url{https://github.com/pkicki/cnp-b}.}
\item A new technique to enforce satisfaction of the boundary constraints, such that the trajectory inferred by the neural network connects precisely two arbitrary robot configurations (positions, velocities, and higher order derivatives when necessary) is proposed and is demonstrated to be effective in both simulations and experiments for two different tasks.
\item Finally, we demonstrate the practical feasibility of our novel planning technique in the Air Hockey hitting task on a real-world Kuka LBR Iiwa 14 robot. This experimental benchmark was previously solved successfully by specialized motion planning algorithms~\cite{liu2021efficient}, while we show the effectiveness of our general planning method under a strict time budget, and we demonstrate its ability to replan motion on-the-fly. These properties allow for precise dynamic hitting motion, outperforming~\cite{liu2021efficient}.
\end{enumerate}


\subsection{Problem Statement}
\label{subsec:problem}
Let $\qt\in \mathbb{R}^n$ be a vector of $n$ generalized coordinates describing a mechanical system at time $t$. Let $\dqt$ and $\ddqt$ be the first and second derivative, i.e. velocity and acceleration of the coordinate vector $\q$. Let the tuple $\traj=\langle \q, \dq, \ddq, T \rangle$ be a trajectory of the system for $t\in\left[0, T\right]$, with $T$ the duration of the given trajectory. In the following, we assume that $\traj(t) = \langle \qt, \dqt,\ddqt \rangle$ is a tuple containing the information of the trajectory $\traj$ at timestep $t$.
We define the constrained planning problem as:
\begin{align}
    \argmin_{\traj}  \quad  & \Loss(\traj) \nonumber\\
    \text{s.t.}  \quad  & \constraint_i(\traj(t), t) = 0 & \forall t, \forall i\in\lbrace 1,\dots, N \rbrace \nonumber\\
    & g_j(\traj(t), t) \leq 0 & \forall t, \forall j\in\lbrace 1,\dots, M \rbrace&,
    \label{eq:problem_formulation}
\end{align}
where $N$ is the number of equality constraints $\constraint_i$, $M$ is the number of inequality constraints $g_j$, and $\Loss$ is an arbitrary loss function describing the task. 
Typically, to solve a motion planning task, it is enough to generate a trajectory that satisfies some locally defined properties, like 
limited joint velocity, torque, or ensuring manipulation in free space, at every time step $t$, and minimize locally defined objectives due to composability.
We can exploit this by transforming the problem into an integral, and describe it by
\begin{equation}
\label{eq:integral_loss}
    \Loss(\traj) = \int_{0}^{T} \stepLoss(\traj(t),t)dt = \int_{0}^{T} \stepLoss(\qt, \dqt,\ddqt, t)dt,
\end{equation}
where $\stepLoss$ is a locally defined loss function.

Instead of specifying the trajectory $\traj$ at every point, it is helpful to use parametric representations, such that the generalized coordinates and their derivatives can be written as $\qt=\bm{f}_{\vtheta}(t)$. Therefore, we can formalize the problem as minimization on the parameters $\vtheta$ instead of searching for $\q$ in the functional space $\mathcal{C}^2$.



\section{Related work}
\label{sec:related_work}

Motion planning is one of the core components of a robot, and it is necessary to achieve the desired degree of autonomy. This importance explains why motion planning is one of the most active research areas in robotics.
Despite these efforts, for many robotic tasks known planning algorithms are inadequate as they cannot handle the task-specific constraints or yield the plan within the specified amount of time. 
Researchers made great progress in fast motion planning in the last two decades, through the development of sampling-based motion
planners such as Probabilistic Roadmaps~\cite{prm} and \gls{rrt}~\cite{rrt}, and their more recent successors like RRT*~\cite{rrtstar}, Bidirectional Fast Marching Trees*~\cite{bfmt}, or Batch Informed Trees*\cite{bitstar_conf}. These methods are not only characterized by probabilistic completeness but also enable planning paths for challenging problems like maneuvering with a car-like vehicle~\cite{NonholonomicRRT} or reaching given poses with two robotic manipulators in a cluttered household environment~\cite{bitstar}.

In parallel with the sampling-based approaches, optimization-based motion planning methods were developed. 
Instead of building a search tree and constructing the solution, these methods start with an initial solution and modify it to minimize cost function subject to constraints. Methods like CHOMP~\cite{chomp}, TrajOpt~\cite{trajOpt} or GPMP2~\cite{mukadam2018gpmp2} showed an ability to solve tasks like avoiding obstacles with PR2 robot~\cite{trajOpt, mukadam2018gpmp2} or planning the walking movements of a quadruped~\cite{chomp}.

Proposing a feasible trajectory for the considered class of planning problems does not mean that a practical solution is reached,
as planning in real-world scenarios often involves additional constraints related to time (e.g. limited time for computing the plan or limited time to complete the motion), or related to the physics of the task (e.g. limited effort of the robot motors or constrained movement of the robot end-effector).
From these constraints, the areas of constrained and kinodynamic motion planning emerged.

\subsection{Constrained motion planning}
Some of the tasks we outsource to robots require constraining the robot's motion to some manifold embedded in the task space, e.g. keeping the cup held by the robot in an upright position or ensuring that the end-effector of a window cleaning robot remains at some constant distance from the window plane. In recent years, two main approaches to this class of problems were developed:
(i) adding task constraints to the motion optimization problem~\cite{constrainedSQP},
(ii) sampling constraint-satisfying configurations and generating constraint-satisfying motions in the sample-based motion planners~\cite{kingston2018sampling,kingston2019exploring}.

While including additional task constraints in motion optimization is conceptually simple, it usually results in a much harder optimization problem to solve, as the kinematics of the robot are usually nonlinear. To overcome the computational burden stemming from these nonlinearities, authors of~\cite{trajOpt} proposed to employ the sequential quadratic programming approach that transforms the nonlinear programming problem into a sequence of its convex approximations around the current solution. In~\cite{constrainedSQP}, this idea was extended to a non-Euclidean setting, by transforming the manifold-constrained trajectory optimization problem into an equivalent problem defined over a space with Euclidean structure.
Another general approach to solve manifold-constrained problems was proposed in~\cite{constrainedCHOMP}, where authors modified the update rule of CHOMP by projecting it on the tangent space of the constraint manifold and adding an offset correction term to move the solution towards the manifold. 
In turn, a recent approach to constrained optimization-based motion planning~\cite{Howell2019ALTRO} tries to achieve faster convergence and numerical robustness by utilizing the augmented Lagrangian optimization using an iterative Linear Quadratic Regulator (iLQR), introduced in~\cite{Li2004iLQR}, and systematically updating the Lagrangian multipliers, which is conceptually similar to our approach to neural network training.  
However, in practice, when we need a solution for a particular problem, exploiting the structure of the constraints may be beneficial. In~\cite{liu2021efficient}, the problem of fast planning for the dynamic motion of a robotic arm with the end-effector constrained to move on a plane was decoupled into two more straightforward optimization problems: planning the Cartesian trajectory on a 2D plane and then planning the trajectory in the joint space. However, this may result in a suboptimal performance due to the predefined Cartesian trajectory. In contrast, in our work, we plan directly in the robot's configuration space, and we show that it is possible to use a general solution to learn how to plan better than using handcrafted optimization.

The adaptation of sampling-based motion planners to solve constrained motion planning problems is not so straightforward. 
Simply rejecting the samples which lie outside of the manifold usually results in a drastically reduced probability of drawing an acceptable sample, also connecting two samples in a manifold-constrained way is non-trivial~\cite{kingston2018sampling,kingston2019exploring}. 
Therefore a number of approaches to this class of problems were proposed, like relaxation, projection, using tangent space or topological atlas.
The relaxation emerged from the assumption that there is some acceptable level of violation of the constraints, and we can relax the surface of the constraint-manifold and give it some non-zero volume to enable sampling~\cite{constrainedSBMPrelaxation1, constrainedSBMPrelaxation2}. However, this approach bypasses the problem by creating a new one -- planning in narrow passages~\cite{szkandera2020}. To mitigate this, \gls{cbirrt} was proposed~\cite{berenson2009manipulation}. This method exploits projections onto the constraint manifold to make sampled and interpolated points satisfy the constraint. In~\cite{wang2015constrained}, authors modified \gls{cbirrt} to avoid the search getting stuck in local minima by exploiting the geometric structure of the constraint manifold.
A significant speed-up over projection-based methods was achieved by Tangent bundle \gls{rrt}~\cite{TBRRT}, which, instead of sampling in the configuration space, proposes to sample in the tangent space of the constraint manifold. Instead of sampling on the tangent space, it is possible to define charts that locally parameterize the manifold and coordinate them by creating a topological atlas~\cite{atlasRRT}. 
In this approach, \gls{rrt} search for the direction of the atlas expansion, and then sampling is performed only in the space defined by the atlas.

Nonetheless, sampling-based constrained motion planning methods usually produce only paths, disregarding the dynamics of the movement that is crucial for planning for dynamic robot movements. 
Therefore, in the following subsection, we will discuss the approaches to kinodynamic motion planning.

\subsection{Kinodynamic motion planning}
Transforming motion optimization problems to the kinodynamic setting can be easily done by including the robot dynamics in the constraints. However, this introduces more nonlinearities to the problems, slows down the solvers, and makes optimization prone to converge to a local minimum unless properly initialized~\cite{Xie2020kinodymanicFactorGraphs}.

In turn, to adopt sampling-based motion planners to the kinodynamic setting, a significant effort has to be made in terms of both the tree Extend procedure and state cost-to-go assessment. One of the first attempts to solve these issues was linearizing the system and using the \gls{lqr} for connecting the states and calculating the cost of this connection~\cite{Perez2012LQRRRT}. To avoid problems stemming from linearization, authors of~\cite{Stoneman2014NLPRRT} proposed to use a gradient-based method to solve the \gls{nl} problem of connecting any two configurations. 
A similar idea was recently presented in~\cite{Primatesta2021MPCRRT}, where \gls{mpc} was used to perform a tree extension.
One of the issues of sampling-based kinodynamic planning is the necessity of sampling high-dimensional state space (due to the inclusion of the velocities in the state). To mitigate this, in~\cite{Zheng2021KinoRRT}, a partial-final-state-free optimal controller was used for connecting the states, such that the dimensionality of the sampling space is divided by two.
A different approach to \gls{rrt}-based planning is to sample controls instead of states~\cite{sst, Cefalo2014TCMP}. This significantly simplifies keeping the constraints satisfied, however, it usually results in relatively slow planning. 
Unfortunately, most of the aforementioned sampling-based kinodynamic motion planning algorithms struggle when they are subjected to kinodynamic and holonomic constraints at the same time~\cite{Bordalba2021atlasKinodynamicRRT}
To address this issue, the adoption of the atlas method~\cite{atlasRRT} with the RRT's Extend procedure using \gls{lqr} was proposed in~\cite{Bordalba2021atlasKinodynamicRRT}. Unfortunately, this algorithm relies strongly on the linearization of the system and requires a significant amount of time to prepare the plan.

\subsection{Learning-based motion planning}
For many real-world robotic applications, conventional planning methods are not fast enough or produce solutions far from optimal. Moreover, tasks performed by the robots, even though they may require solving different planning problems, seem to reveal some similarities.
Therefore, to obtain plans of better quality within tighter time bounds, learning-based approaches to improve robot motion planning were proposed. These types of methods originated from~\cite{Berenson2012remember}, in which, the conventional planner was used to solve the problems, however, generated plans were not only executed but also saved in the library of solved problems. After gathering plans, they can be reused to solve new but similar problems in a shorter time. 

Most of the learning-based motion planning methods build upon the algorithms from \gls{rrt} family to use their guaranties of probabilistic completeness. Some of these methods try to modify the sampling distribution based on the experience and the representation of the task. In~\cite{Zhang2018distribution} neural network decides whether the sampled state should be rejected or not, whereas in~\cite{Huh2018rrtq} neural network is used to predict the state-action value function to choose the best nodes to expand. Authors of~\cite{Molina2020learnAndLink} used a \gls{cnn} to predict some crucial regions of the environment to increase the sampling in them. At the same time, in~\cite{Cheng2020distribution}, a probability distribution is directly inferred by the neural network in the form of a heatmap. In~\cite{Quershi2021MPNet}, the sampling distribution is encoded in a stochastic neural network (due to dropout used during inference), which is trained to predict the next state towards the goal.

An adaptation of biasing the sampling distribution to comply with a manifold of constraints was presented in~\cite{Lembono2021constrainedDistribution}, where adversarial training was used to learn how to generate data on the constraint manifold. An architecture consisting of a generator and discriminator trained jointly was also used in~\cite{CoMPNetX}. This work builds upon~\cite{Quershi2021MPNet} and uses a generator to predict the next state, but extends it with a discriminator, which is used to predict deviation from the constraint manifold and to project the predictions on it. 

Learning-based approaches are also used to improve the feasibility of solving kinodynamic motion planning problems. Authors of~\cite{Wolfslag2018KMPLearnSteering} emphasized that solving a two-point boundary value problem for a nonlinear dynamical system is NP-hard and proposed to learn a distance metric and steering function to connect two nodes of \gls{rrt}, based on the examples generated using indirect optimal control. In turn, in~\cite{Atreya2022KMPLearnSteering} the learning of control policy is done indirectly by learning the state trajectory of some optimal motion planner and then using inverse dynamics of the system to determine controls. 
A lazy approach to limit the number of computationally expensive \gls{nl} solver calls was presented in~\cite{Yavari2019KMPLazy}, where neural network, instead of replacing the solver, is trained to predict which nodes of a \gls{rrt} are steerable to, and what will be the cost of steering. Contrary to these methods, authors of~\cite{li2021mpcmpnet} propose to use a \gls{mpc} extensively to solve local \gls{nl} problems and learn only how to determine the next state in a way towards the goal using the experience gained by mimicking demonstration trajectories.

In contrast to these methods, our proposed approach offers an extremely fast way of finding a near-optimal trajectory that is able to solve constrained kinodynamic motion planning problems. Moreover, \gls{ours} does not require the demonstration trajectories, as it learns from its own experience similarly to~\cite{kicki2021learning}. Thus, its performance is not upper-bounded by the quality of the demonstrations. Furthermore, our proposed approach is able to smoothly replan the motion on-the-fly, which is not possible with the use of state-of-the-art constrained kinodynamic motion planning algorithms.

\section{PROPOSED SOLUTION}
In this section, we will first describe the general framework of our proposed solution to the problem defined in \eqref{eq:problem_formulation}. Then we will show how to use the structure of the proposed approach to solve the problem of planning a trajectory where the objective is to reach a specific end configuration in the shortest possible amount of time. Finally, we will show some extensions to \gls{ours} to solve a real planning problem. Namely, we will focus on moving a heavy object and hitting a puck with a mallet using a 7dof manipulator robot.

\subsection{Constrained planning}
\label{sec:constrained_planning}
The original problem formulation, shown in~\eqref{eq:problem_formulation}, is a complex constrained optimization problem: directly using this formulation makes finding a solution hard and time-consuming, particularly in the presence of complex equality constraints. 
Indeed, while an exact solution may exist, it may be hard to perfectly match the constraints, due to the use of parametrized trajectory and numerical errors.
To efficiently solve the constrained planning problem, we reformulate the original optimization, introducing the notion of constraint manifold. This reformulation transforms the constrained optimization problem into an unconstrained one by modifying the original loss such that the solutions lie close to the constraint manifold. In our approach, we learn the loss during the optimization process by considering the manifold metric and a violation budget for each constraint.

We base our reformulation on the following steps: 
\begin{enumerate}[label=\roman*.]
    \item defining the constrained manifold as the zero-level curve of an implicit function, unifying equality and inequality constraints,
    \item relaxing the original problem by imposing that all solutions should lie in the vicinity of the constraint manifold,
    \item transforming this problem into an unconstrained optimization problem by learning the metric of the constraint manifold.
\end{enumerate}

\subsubsection{Defining the constraint manifold}
Our first step towards the definition of the constraint manifold is to eliminate inequality constraints from~\eqref{eq:problem_formulation}, by introducing slack variables $\slack_j$ and $M$ new equality constraints $\constraint_{N+j}(\q, \dq, \ddq, t, \slack_j) = g_{j}(\q, \dq, \ddq, t) + \slack_j^2$. Thus, we obtain the following formulation
\begin{align}
    \argmin_{\traj}  \quad  & \Loss(\traj) \nonumber\\
    \text{s.t.}  \quad  & \constraint_i(\traj(t), t, \slack_i) = 0 & \forall t, \forall i\in\lbrace 1,  \dots, N+M\rbrace,
    \label{eq:problem_formulation_equality}
\end{align}
where set $\lbrace(\traj(t), \slack) \,|\, \constraint_i(\traj(t), t, \slack_i) = 0, \quad \forall i\in\lbrace 1,  \dots, N+M\rbrace\rbrace$ defines the constraints manifold $\manifold_\slack$. To drop the dependency of the manifold $\manifold_\slack$ from the vector of slack variables $\Slack\in\mathbb{R}^{M}$, we first introduce the manifold loss function 
\begin{align}
\label{eq:manifold_loss}
    \manifoldLoss(\traj(t), t, \Slack) &= \|\Constraint(\traj(t), t, \Slack)\|^2 \nonumber\\
    & = \sum_{i=0}^m c_i(\traj(t), t, \slack_i)^2  = \sum_{i=0}^m \manifoldLoss^i(\traj(t), t, \slack_i),
\end{align}
where $\|\cdot\|$ denotes a L2 norm, such that the constraint manifold $\manifold$ will be defined by its 0-level set. 
For $i$-th relaxed inequality constraint
the manifold loss function is defined by
\begin{align}
     \manifoldLoss^i(\traj(t), t, \slack_i) & = c_i(\traj(t), t, \slack_i)^2 = \left(\constraint_i(\traj(t), t) + \slack_i^2\right)^2 \nonumber \\
          & = \constraint_i(\traj(t), t)^2 + 2\constraint_i(\traj(t), t)\slack_i^2 + \slack_i^4,
\end{align}
therefore, to drop the dependency of $\slack_i$, we redefine our loss for inequality constraint as follows
\begin{equation}
     \manifoldLoss^i(\traj(t), t) = \min_{\slack}\manifoldLoss^i(\traj(t), t, \slack_i).
\end{equation}
We can find a value that minimizes $\manifoldLoss^i(\traj(t), \slack_i)$ by taking the derivative w.r.t. the slack variable and setting it to zero
\begin{equation}
    \dfrac{d}{d\slack}\manifoldLoss^i(\traj(t), t, \slack_i) = 4\constraint_i(\traj(t), t)\slack_i + 4\slack_i^3 = 0,
\end{equation}
thus, we obtain $\slack_i = 0 \vee \slack_i = \pm \sqrt{-\constraint_i(\traj(t), t)}$, where the second solution exists only for $\constraint_i(\traj(t), t) < 0$. Plugging back the obtained stationary points in the solution, by basic reasoning of the obtained values and conditions, we obtain our inequality loss
\begin{equation}
     \manifoldLoss^i(\traj(t), t) = \begin{cases} 
     \constraint_i(\traj(t), t)^2 & \constraint_i(\traj(t), t) > 0\\
     0 & \text{otherwise}. \\
     \end{cases}
\end{equation}
This loss can be computed compactly, in an equivalent form, as
\begin{equation}
\label{eq:inequality_relu}
     \manifoldLoss^i(\traj(t), t) = \left(\max \left(\constraint_i(\traj(t), t), 0\right)\right)^2,
\end{equation}
which is continuous and differentiable everywhere: notice that in 0, the derivative is 0.
As a result, we obtain a new definition of the constraint manifold 
\begin{align}
    \manifold \triangleq \lbrace\traj(t) \,|\, \manifoldLoss^i(\traj(t), t) = 0, \quad \forall t, \forall i\in\lbrace 1,  \dots, N+M\rbrace\rbrace.
\end{align}

\subsubsection{Approximated optimization problem} Our newly defined manifold loss expresses the distance of a given trajectory $\traj$ from the manifold $\manifold$.
Ideally, we could use the $\manifoldLoss(\traj, t) = 0$ as a single constraint of our optimization problem. However, numerical issues and trajectory parametrization may cause small unavoidable errors. Furthermore, often the task loss $\Loss$ and manifold constraints $\manifoldLoss^i$ will counteract each other, making this optimization particularly difficult and prone to constraint violations unbalances. Indeed, the optimization may favor reducing some constraints violation at the expense of others.
When considering real-world tasks, it is often not crucial to have zero constraint violation as long as it is below an acceptable threshold.
Thus, we simplify the problem by introducing an acceptable level of constraint violation. For simplicity, we bound the elements of the non-negative valued manifold loss function (see \eqref{eq:manifold_loss} and \eqref{eq:inequality_relu}) i.e. $\manifoldLoss^i \leq \bLMi$, where $\bLMi = \bar{\constraint}^2_i$ is a square of the desired acceptable constraint violation level $\bar{\constraint}_i$. Thus, we can write the following optimization problem 
\begin{align}
    \argmin_{\traj}  \quad  & \Loss(\traj) \nonumber\\
    \text{s.t.}  \quad  & \manifoldLoss^i(\traj(t), t) \leq \bLMi & \forall t, \forall i\in\lbrace 1,  \dots, N+M\rbrace.
    \label{eq:problem_formulation_epsilon}
\end{align}
Now, we can transform~\eqref{eq:problem_formulation_epsilon} into the canonical form
\begin{align}
    \argmin_{\traj}  \quad  & \Loss(\traj) \nonumber\\
    \text{s.t.}  \quad  & \frac{\manifoldLoss^i(\traj(t))}{\bLMi} - 1 \leq 0 & \forall t, \forall i\in\lbrace 1,  \dots, N+M\rbrace.
    \label{eq:problem_formulation_canonical}
\end{align}

One possible way to solve~\eqref{eq:problem_formulation_canonical} is by applying a Lagrange relaxation technique \cite{bertsekas_largrange}, framing it as an unconstrained optimization of the following function
\begin{align}
    \argmin_{\traj} \max_{\Lm} \quad &  L(\traj, \Lm) = \Loss(\traj) + \sum_{i=1}^{N+M}\lm_i\pare{\frac{\manifoldLoss^i(\traj)}{\bLMi}-1} \nonumber \\
    \text{s.t.} \quad  & \lambda \geq 0,
    \label{eq:problem_formulation_minmax}
\end{align}
where $L(\traj(t), \Lm)$ is the Lagrangian and $\Lm$ is a vector of non-negative Lagrange multipliers. 

\subsubsection{Loss function learning}
Unfortunately, \eqref{eq:problem_formulation_minmax} is not a practical optimization problem, leading to ineffective learning and convergence to local optima.  
We propose instead an approximate solution, inspired by~\cite{stooke2020responsive}, that learns the Lagrangian multipliers. In our approach, we decouple the min-max problem into interleaving minimization of~\eqref{eq:problem_formulation_minmax} w.r.t $\traj$ and updating the values of the multipliers $\Lm$. We frame the update of the multipliers as the process of learning a metric of our constraint manifold, weighting each constraint by how much it is important to solve the final task in a way that is feasible in practice.

First, we remove the maximization w.r.t $\Lm$ from~\eqref{eq:problem_formulation_minmax} and write the following minimization problem
\begin{equation}
\label{eq:problem_formulation_min}
    \argmin_{\traj(t)} \Loss + \Constraint^T \Lambda \Constraint,
\end{equation}
where matrix $\Lambda$ is a metric of the manifold $\manifold$ defined by
\begin{equation}
    \Lambda = \diag\pare{\frac{\lm_1}{\bLM_1}, \frac{\lm_2}{\bLM_2}, \ldots, \frac{\lm_{N+M}}{\bLM_{N+M}}}.
\end{equation}
Note that we drop the constant 1 from the constraints, as we are not solving a min-max optimization problem, and this term has no effect on the minimization part.

Second, to simplify this metric and to maintain the positiveness of all elements of $\Lm$ we propose the following substitution $\alpha_i = \log \frac{\lm_i}{\bLMi}$. Thus, we write $\Lambda$ as
\begin{equation}
    \Lambda = \diag e^{\alphas} = \diag\pare{e^{\alpha_1}, e^{\alpha_2}, \ldots, e^{\alpha_{N+M}}}, 
\end{equation}
where $\alphas$ is a vector of real-valued parameters defining the manifold metric. We introduce now a new loss, the manifold loss under the metric $\Lambda$
\begin{equation}
    \manifoldMetricLoss = \Constraint^T \Lambda \Constraint.
\end{equation}
As a result we can rewrite \eqref{eq:problem_formulation_min} into 
\begin{equation}
    \argmin_{\traj} \Loss(\traj) + \manifoldMetricLoss(\traj).
    \label{eq:problem_formulation_min_compact}
\end{equation}

Third, we introduce the metric learning approach.
While we can straightforwardly optimize~\eqref{eq:problem_formulation_min_compact} w.r.t. $\traj$ using any gradient optimizer, we need to derive a learning rule for the vector $\alphas$.
Let $\cLMi \triangleq \Constraint^T_{\neg i}\Lambda_{\neg i}\Constraint_{\neg i}$ be the complement of the $i$-th manifold loss element,
where index $\neg i$ means all elements except $i$-th element. Using this notation, we analyze how the $\alpha_i$ should change between $k$-th and $k+1$-th iterations so as not to surpass the desired level of constraint violation $\bLMi$, i.e.
\begin{equation}
  \Loss^{(k)} + {\cLMi}^{(k)} + e^{\alpha^{(k)}_i} \manifoldLoss^i = \Loss^{(k+1)} + {\cLMi}^{(k+1)} + e^{\balpha_i} \bLMi.
\end{equation}
Assuming that the changes of $\cLMi$ and $\Loss$ are small, e.g. by choosing a small learning rate, we obtain
\begin{equation}
   e^{\alpha^{(k)}_i} \manifoldLoss^i = e^{\balpha_i} \bLMi \Rightarrow \balpha_i = \alpha^{(k)}_i + \log\pare{\frac{\manifoldLoss^i}{\bLMi}}.
\end{equation}
Finally, we can define our update rule for $\alpha_i$ as a small step towards the desired value $\balpha_i$ i.e.
\begin{equation}
    \Delta \alpha_i = \gamma (\balpha_i - \alpha_i) = \gamma \log\pare{\frac{\manifoldLoss^i}{\bLMi}},
\end{equation}
with the learning rate $\gamma \in \Rplus$.

\subsection{Trajectory Parameterization}
\label{sec:trajectory_parametrization}
To represent the solution trajectory $\traj$ we use a decoupled representation made up of time $t = \tfroms$ and path $\Paths = \q(\tfroms)$, both defined as functions of the phase variable $\s \in [0; 1]$, where $\tfroms$ is monotonically increasing, $\psi(0) = 0$, and $\psi(1) = T$, where $T$ is duration of the trajectory. Therefore, $\qt$ can be defined by
\begin{equation}
    \qt \triangleq \q(\tfroms) = \Paths.
\end{equation}
The first derivative of $\q$ w.r.t. time can be defined by
\begin{equation}
    \dqt = \dPaths \dsdt = \dPaths \rates,
\end{equation}
where $\rates \triangleq \pare{\frac{\dt}{\ds}}^{-1}(s)$ represents the inverse rate of change of the time $t$ w.r.t. phase variable $\s$.
Thus, the second derivative of $\q$ w.r.t. time can be defined by
\begin{equation}
    \begin{split}
    \ddqt = &\overdt\pare{\dPaths} \rates + \dPaths \overdt \rates=\\
    & \ddPaths \pare{\rates}^2 + \dPaths \drates \rates.
\end{split}
\end{equation}
Therefore, to fully represent trajectory $\traj$, we need to know path $\Paths$ and the inverse rate of change of time $\rates$, and their derivatives w.r.t. phase variable $s$. In this paper, we represent $\Paths$ and $\rates$ as B-splines, as this parametrization is rich enough to generate complex trajectories and allow us to easily compute their derivatives, which is crucial to ensure the satisfaction of the $\dqt$ and $\ddqt$ limits.

\subsection{Trajectory definition}
To define the trajectory, we must determine the control points $\qControlPoints, \tControlPoints$ of both $\Paths$ and $\rates$ B-splines. Most of these control points are predicted using the neural network, however, part of them can be directly computed based on the task definition.

\subsubsection{Boundary conditions}
The majority of the practical motion planning problems impose some boundary constraints on their solutions. Typically, at least the initial configuration is defined. However, for practical problems, it is beneficial to consider also initial velocity and accelerations, to ensure smooth movement from the given initial state. Similarly, some tasks require accurate reaching of some desired state and velocity. These constraints can be summarized as 
\begin{equation}
\label{eq:boundary_constraints}
\begin{cases}
    \q(0) = \q_0,\quad \dq(0) = \dq_0,\quad \ddq(0) = \ddq_0,\\
    \q(1) = \q_d,\quad \dq(1) = \dq_d\,
\end{cases}
\end{equation}
where $\q_0, \dq_0, \ddq_0, \q_d, \dq_d$ are the initial and desired configurations and their derivatives given by the task definition.
By using B-spline representations of $\Paths$ and $\rates$, we can impose these boundary conditions directly onto the solution of the considered planning problem. Technical details of this procedure are described in Appendix \ref{ap:bspline}. As a result, the first three and last two robot's configuration control points $\qControlPoints$ may be determined based on the boundary constraints of the problem \eqref{eq:boundary_constraints} instead of using the neural network outputs. 

\subsubsection{Neural network}
\begin{figure}[bt]
    \centering
    \includegraphics[width=\linewidth]{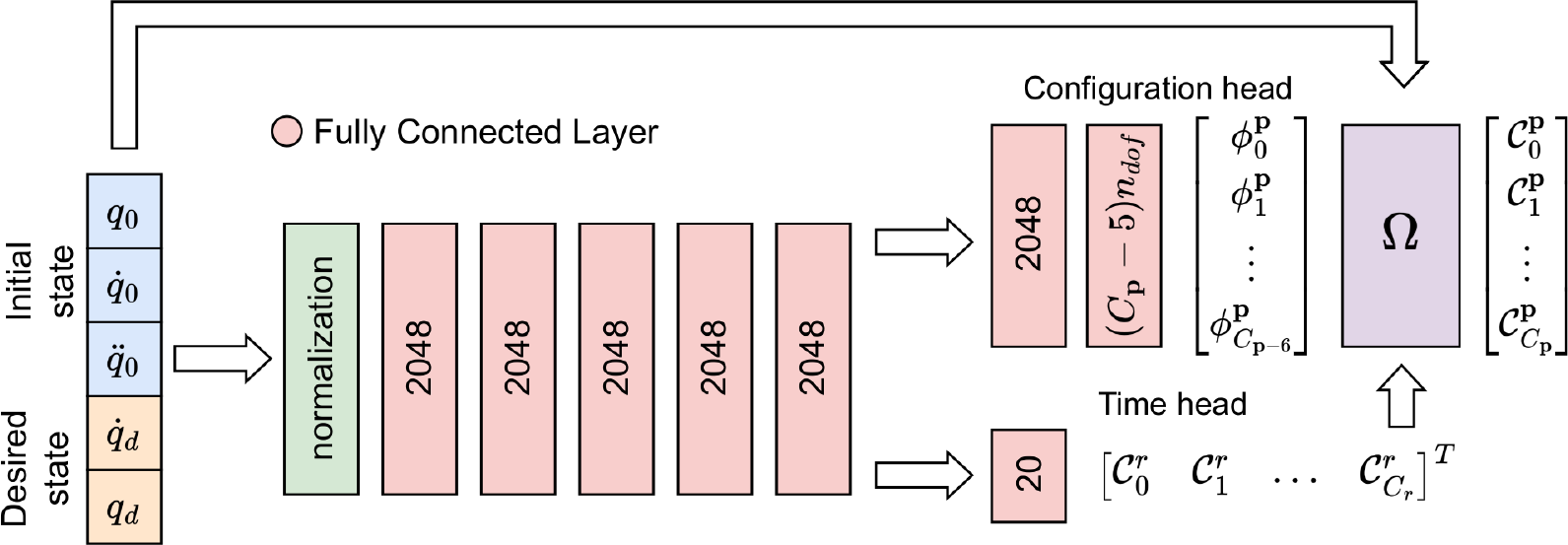}
    \caption{Architecture of the neural network used to determine $\Paths$ and $\rates$ B-spline control points. The $\Omega$ Block computes the control points of the configuration B-spline $\qControlPoints$ using \eqref{eq:controlPoints} and \eqref{eq:configuration_control_points}.}
    \label{fig:nn}
\end{figure}

While boundary control points can be computed analytically, the rest of the control points must be determined using the learning system based on previous experience. Our proposed neural network (Fig.~\ref{fig:nn}) takes as input the normalized initial and desired configurations $\qz, \qd$, velocities $\dqz, \dqd$ and initial joint accelerations $\ddqz$, and outputs the control points $\tControlPoints$ of the $\rates$ B-spline, and some parameters~$\bm{\NNPathOutput}$, which can be used to compute the control points of the configuration B-spline $\qControlPoints$.
Activation functions for all layers are $\tanh$, except the layer in the \textit{Time head}, which is an exponential function (to ensure the positiveness for all control points of the $\rates$ B-spline). 

To compute control points of the configuration B-spline, we utilize the outputs $\bm{\NNPathOutput}$ of the \textit{Configuration head} oriented with the boundary control points of the $\Paths$ B-spline. For the case for which the initial state is defined with the configuration and its derivatives up to $N_i$-th (exclusively), and the desired state with the configuration and its derivatives up to $N_d$-th (exclusively), we can define inner configuration control points by
\begin{align}
\label{eq:configuration_control_points}
    \qControlPoints_{i+N_i} = \pare{\qControlPoints_2 \pare{1 - \frac{i}{\numQControlPoints - B}} + \qControlPoints_{\numQControlPoints - N_d} \frac{i}{\numQControlPoints - B}} + \pi\NNPathOutput_{i}\nonumber
    \\\quad\text{for}\quad 0 \leq i \leq \numQControlPoints - B,
\end{align}
where $\numQControlPoints$ is the total number of configuration control points $\qControlPoints$, $B=N_i+N_d+1$, and $\NNPathOutput_{i}$ is the $i$-th output from the configuration head.

\subsection{Loss functions}
The loss function is one of the most crucial parts of every learning system. In the proposed solution, we consider two types of losses i.e. task loss $\Loss$ and manifold loss $\manifoldLoss$, however, we include both of them into the integral formulation \eqref{eq:integral_loss}, such that the resultant loss function $\nnLoss$ can be defined by
{\small
\begin{equation}
    \nnLoss(\traj) = \int_0^T \stepNNLoss(\traj(t), t) dt = \int_0^T \pare{\stepLoss(\traj(t), t) + \stepManifoldLoss(\traj(t), t)} dt,
\end{equation}}
where $\stepLoss(\traj(t), t)$ and $\stepManifoldLoss(\traj(t), t)$ are locally defined task and manifold losses at time $t$.
To simplify the optimization process, we drop the explicit dependency of the loss $\nnLoss$ on the time $t$ and trajectory duration $T$, by the following change of variables
\begin{equation}
    \nnLoss(\traj) = \int_0^1 \stepNNLoss(\traj(t), t) \frac{\dt}{\ds} \ds =  \int_0^1 \stepNNLoss(\traj(s), s) \rates^{-1} \ds.
\end{equation}

Therefore, to define the resultant loss function $\nnLoss$, we must provide the formulas for the step losses $\stepLoss(\traj(s), s)$ and $\stepManifoldLoss(\traj(s), s)$. 
In our experiments, to encourage the neural network to generate minimal-time (i.e. fast) trajectories, we define the basic task step loss as $\stepLoss(\traj(s)) = 1$. However, we can optimize any other quantity, like effort, jerk, end-effector position tracking error, centrifugal forces, or a weighted sum of these.

To define step manifold loss $\stepManifoldLoss(\traj(s), s)$ we need to define its components $\stepManifoldLossi(\traj(s), s)$, which corresponds to all constraints imposed in the given problem. 
In our experiments, we consider constraints stemming from velocity $\dqLimit$, acceleration $\ddqLimit$, which associated losses can be defined by
\begin{equation}
    \stepManifoldLoss^{\dq}(s) = \sum_{i=1}^{\ndof} \huber(\relu(\abs{\dqi(s)} - \dqLimiti)),
\end{equation}
\begin{equation}
    \stepManifoldLoss^{\ddq}(s) = \sum_{i=1}^{\ndof} \huber(\relu(\abs{\ddqi(s)} - \ddqLimiti)).
\end{equation}
Similarly, we can define loss for torque limits $\torqueLimit$
\begin{equation}
    \stepManifoldLoss^{\torque}(s) = \sum_{i=1}^{\ndof} \huber(\relu(\abs{\torquei(s)} - \torqueLimiti)),
\end{equation}
where $\torque(s) = \ID(\q(s), \dq(s), \ddq(s))$ can be easily computed using inverse dynamics algorithm~\cite{carpentier2019pinocchio, struz2017iiwaparams}.
Moreover, in a similar manner, we can define losses that utilize other functions of the robot configurations, like the forward kinematics. Therefore, it is easy to define some sample geometrical constraints
\begin{equation}
    \stepManifoldLoss^{\taskSpaceManifold}(s) = \huber(d(\FK(\q(s)), \taskSpaceManifold)),
\end{equation}
where $d(\FK(\q(s)), \taskSpaceManifold)$ is a distance between the robot's links configurations and some manifold $\taskSpaceManifold$ in the task space.

Obviously, the aforementioned losses are only examples. Using this framework, we can come up with much more complicated and sophisticated loss functions, however, the presented loss functions are enough to learn how to perform challenging practical tasks considered in this paper. Note, that all of these losses are differentiable functions of the trajectory $\traj$, which enables us to optimize the neural network weights directly.

\section{EXPERIMENTAL EVALUATION}

\begin{figure*}[t!]
    \centering
    \includegraphics[height=0.25\textwidth]{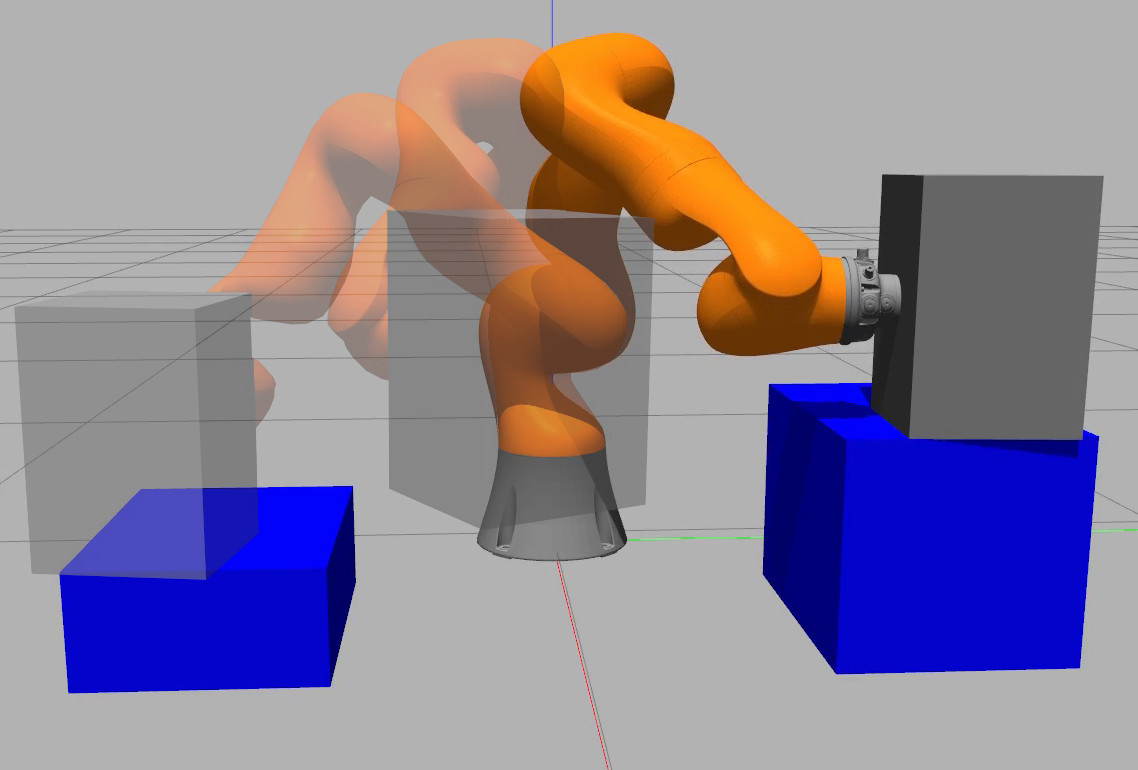}
    \hspace{1.5cm}
    \includegraphics[height=0.25\textwidth]{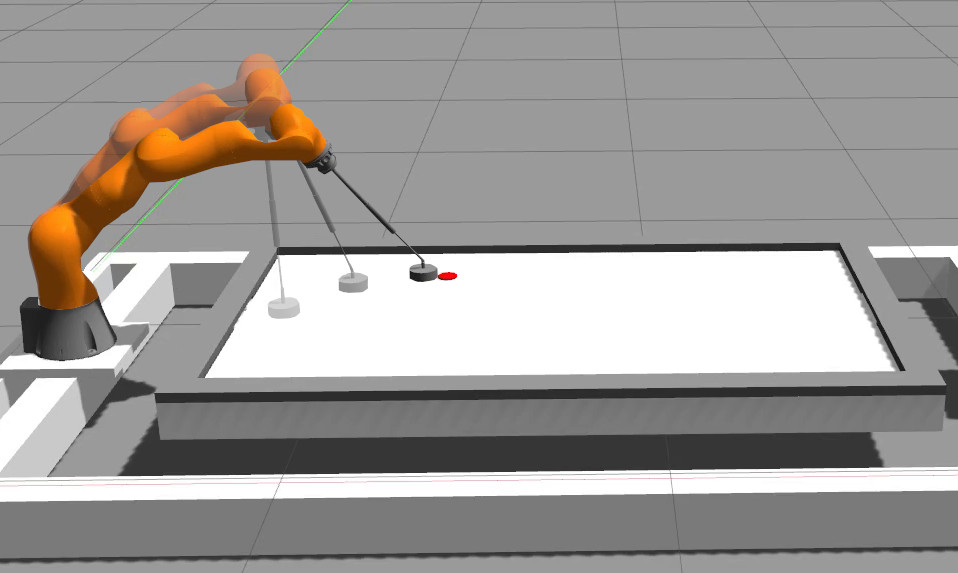}
    \caption{Two tasks considered in this paper: moving a heavy vertically oriented object between two tables (left) and high-speed Air Hockey hitting (right).}
    \label{fig:tasks}
\end{figure*}

To evaluate the proposed constrained neural motion planning framework, we introduce two challenging motion planning tasks, both considering planning for the Kuka LBR Iiwa 14 robotic manipulator: moving a heavy (of a weight close to the robot's payload limit) vertically oriented object between two tables and high-speed hitting in the game of robotic Air Hockey (see Fig.~\ref{fig:tasks}).
The baseline, to which we compare the performance of \gls{ours} in both tasks is constituted by four algorithms that span the space of different solutions to the considered problem:
\begin{itemize}
    \item TrajOpt~\cite{trajOpt} -- a motion optimization algorithm that utilizes \gls{slsqp},
    \item \gls{cbirrt} -- a sampling-based path planning algorithm that uses RRTConnect with the projection of sampled points onto the constraint manifold,
    \item \gls{sst} -- a sampling-based motion planning algorithm which builds a sparse tree of robot configurations and extends them using random controls,
    \item \gls{mpcmpnet} -- a sampling-based motion planning algorithm, which uses a neural network to determine the next node in a search tree and \gls{cem} to steer towards this configuration.
\end{itemize}
In the Air Hockey hitting task we extend the baseline by the \gls{aqp} method, which is the current state-of-the-art in this area -- an algorithm developed specifically to solve the problem of planning on the constraint manifold of the Air Hockey game.

Both the baselines and our method were evaluated in a simulation environment developed using ROS1 and Gazebo.
The experiments in Air Hockey hitting were possible thanks to the in-house constructed physical setup with a Kuka LBR Iiwa 14 robot~\cite{liu2021efficient}.
For the experiments in simulation, we used an Intel Core i7-9750H CPU, while the real robot experiment uses AMD Ryzen 9 3900x CPU.

\subsection{Moving a heavy vertically-oriented object}

\begin{figure*}[b!]
    \centering
    \includegraphics[width=\linewidth]{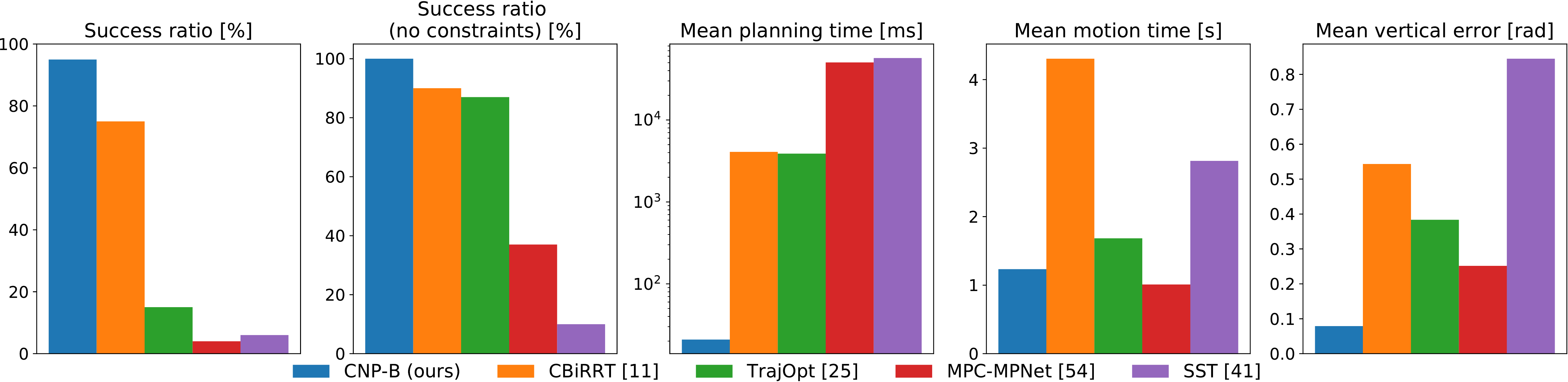}
    \caption{Planners statistics on the task of rapid movement of a heavy object with orientation constraints and collision avoidance.}
    \label{fig:kino_stats}
\end{figure*}

\subsubsection{Task description} In this task, the goal is to quickly move a heavy cuboid (12 kg) between random positions on the left and right blue boxes using Kuka LBR Iiwa 14. 
The task requires planning a joint trajectory between two random configurations, minimizing movement time, and satisfying joints’ velocity, acceleration, and torque constraints. 
Moreover, the robot's trajectory has to ensure that both the robot and the handled object will not collide with the environment and that the object will be oriented vertically throughout the whole movement.

\subsubsection{Dataset and method adjustments} To learn how to solve the task, we generated a dataset of 26400 planning problems of this kind, split into training (24000) and validation (2400) subsets.
In the dataset, we randomize both the initial and desired position of the object and the initial and desired robot's configuration. Both initial and desired velocities are set to 0.
Moreover, we add three additional loss terms stemming from the constraints imposed by the task definition i.e. vertical orientation loss, robot collision loss, and object collision loss. The mathematical definitions of these losses, together with all the parameters of this experiment, can be found in Appendix~\ref{ap:kinodynamic_task}.

\subsubsection{Quantitative comparison with state-of-the-art} 
To evaluate the proposed method, we compared it with several state-of-the-art motion planners. All planners were asked to plan the motions that solve 100 randomly generated tasks, and we executed these plans in simulation.
The results of this experiment are presented in Fig.~\ref{fig:kino_stats}. The first two plots show that our planner reaches the goal in all scenarios, and in 95\% does not violate any of the constraints. In contrast, comparable results are obtained only by \gls{cbirrt}~\cite{berenson2009manipulation}, which reaches the goal in 88\% of cases, and only 74 plans out of 100 are valid. However, this method plans motions that are on average two times longer, and it takes over 250 times longer to compute this solution.
The result of the \gls{mpcmpnet}~\cite{li2021mpcmpnet} deserves special attention, as it is the state-of-the-art learning-based solution for kinodynamic motion planning. We trained it using the plans generated by our planner, however, it was unable even to come close to the results achieved by the TrajOpt~\cite{trajOpt} and \gls{cbirrt}~\cite{berenson2009manipulation}, not to mention our proposed planner. Surprisingly, it outperforms all solutions in terms of the mean motion time, however, it is the result of generating plans that are too short to reach the goal. 
The last plot shows the error of maintaining the object in an upward position, which we computed as an integral along the trajectory of the sum of the absolute values of roll and pitch angles. The smallest deviation from the orientation constraint is achieved by our proposed solution, while the highest violations are generated by executing the trajectories planned using \gls{sst}~\cite{sst}.

\subsection{Planning high-speed hitting in simulated Air Hockey game}

\begin{figure*}[t]
    \centering
    \includegraphics[width=\linewidth]{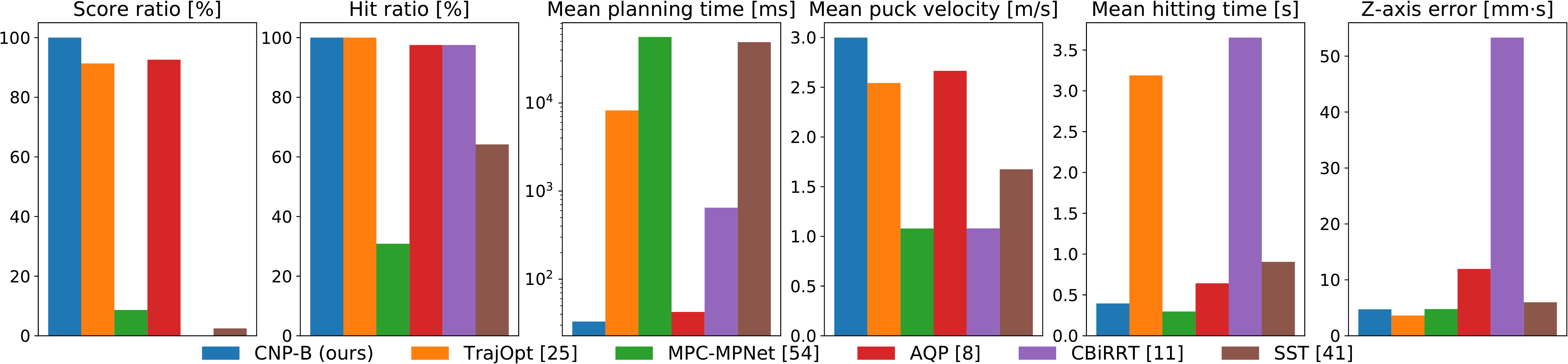}
    \caption{Planners statistics on the task of hitting in the simulated robotic Air Hockey.}
    \label{fig:ah_sim_stats}
\end{figure*}

\subsubsection{Task description}

In this task, the goal is to score a goal in the game of robotic Air Hockey from a steady-still puck i.e. to move the Kuka LBR Iiwa 14 robot handling the mallet from some predefined initial configuration, such that it will reach the puck position with the velocity vector pointing towards the middle of the goal.
Robot desired configurations and joint velocities are determined using the optimization algorithm proposed in~\cite{liu2021efficient} for a given puck position and velocity direction.
Moreover, the planner should generate a joint trajectory
minimizing movement time and satisfying joints' velocity, acceleration, and torque constraints.

\subsubsection{Dataset and method adjustments}
\label{sec:sim_dataset}
To learn the task, we generated a dataset of 19800 planning problems of this kind, split into 2 subsets: training (18000) and validation (1800), while the test set is defined separately. The robot's initial configuration is randomly drawn in a neighborhood of a base configuration, such that the mallet is located in a 10$\times$10 cm box, and its initial velocity and acceleration are set to 0. The puck position is also random, and the hitting direction is computed to approximately point towards the goal.
Moreover, we add an additional constraint manifold loss term that penalizes the displacement of the robot end-effector from the table surface, and a task loss term that penalizes the high centrifugal forces at the end-effector, to reduce the trajectory tracking errors.
The mathematical definition of these losses, together with all parameters of this experiment, can be found in Appendix~\ref{ap:ah_sim_task}.

\subsubsection{Quantitative comparison with state-of-the-art} We compared our approach with state-of-the-art motion planning algorithms on the set of 81 hitting scenarios of a puck being located on a 9$\times$9 grid, which were not present in the training and validation sets. 
One of the key challenges of this task is that the goal configuration has to be reached with a given high velocity in order to score the goal. It is hard to plan the trajectory that satisfies this kind of constraint, especially using sampling-based motion planners. Therefore for \gls{cbirrt}, \gls{mpcmpnet} and \gls{sst} algorithms, we simplified the task, such that the goal was to at least reach the puck position i.e. we computed the distance function only for positional coordinates.

The results of this comparison are presented in Fig.~\ref{fig:ah_sim_stats}. Even though the task for \gls{mpcmpnet} and \gls{sst} planners is simplified, they cannot plan trajectories within a reasonable time. The only sampling-based algorithm which is able to reach the target is \gls{cbirrt}, which produces plans very hard to follow (see z-axis error chart in~Fig.\ref{fig:ah_sim_stats}). 
Far better performance is achieved by optimization-based planners (TrajOpt and \gls{aqp}), which almost always hit the puck and score,  respectively in 91.35\% and 92.59\% of scenarios. The limitation of TrajOpt is that it needs nearly 10s on average to compute the plan, whereas the \gls{aqp} mean planning time is 42ms.
A similar planning time scale is achieved only by our proposed algorithm, which plans within 33ms. Moreover, our solution achieves a 100\% of scoring ratio, plans the fastest trajectories (over 60\% faster than \gls{aqp}), obtains the highest hitting velocities, and despite this, it generates the trajectories that are possible to follow with the accumulated $z$-axis deviation smaller than 5mm$\cdot$s.
Interestingly, \gls{aqp} is a state-of-the-art solution tailored specifically to planning for robotic Air Hockey, and yet its performance is dominated in terms of all considered criteria by the solution trained on automatically generated data using our proposed general framework.

\subsubsection{Qualitative results for replanning}
\begin{figure*}[t]
    \centering
    \includegraphics[width=\linewidth]{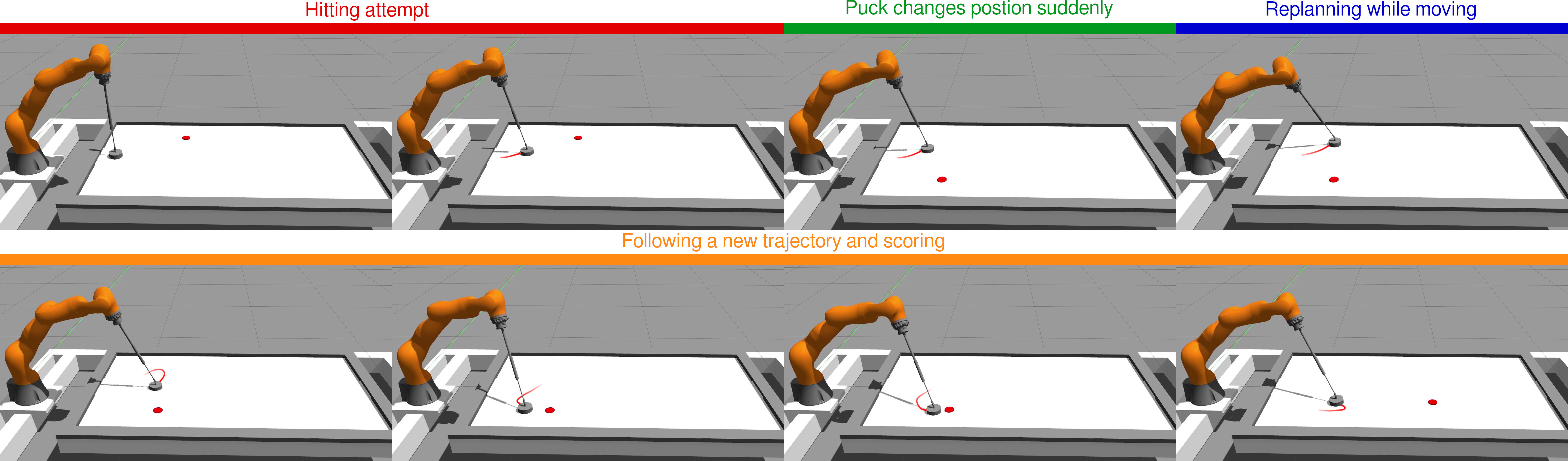}
    \caption{Sequence of frames form the replanning scenario. The robot starts the hitting motion with the puck located in the upper part of the table, but after 300ms puck is moved to the lower part. In response to this, \gls{ours} immediately replans the trajectory and scores the goal.}
    \label{fig:ah_sim_replanning}
\end{figure*}

So far, we have quantified how the proposed solution compares to state-of-the-art solutions. However, due to the short and deterministic planning time, and the ability to satisfy boundary conditions, our proposed approach allows for solving tasks, that are impossible to solve using state-of-the-art planners i.e. replanning on the fly.
We now consider a situation when the robot is performing some plan, and in the meantime, the goal changes e.g. puck's expected position or desired hitting direction has changed. For this type of task, the planning time of almost all state-of-the-art methods is too long to react. Moreover, typical motion planning methods do not give any guarantee about the maximal planning time. Unlike these classical approaches, our solution needs a small constant amount of computation to plan the motion. 
Therefore, we can predict a robot configuration located forward in time along the current trajectory, and plan from this configuration, taking into account the smoothness of the motion and continuity of actuation, by imposing the boundary conditions on the planned motion.

To learn how to plan for a moving robot, we prepared a more general (compared to the one introduced in Section~\ref{sec:sim_dataset}) dataset of hitting scenarios, which includes hitting from many different robot configurations in the accessible space, with random initial velocities and accelerations, and random desired hitting directions and velocities. A more detailed description of the used datasets can be found in Appendix~\ref{ap:datasets}.

In Fig.~\ref{fig:ah_sim_replanning} we show a sequence of frames from the scenario where the robot tries to hit the puck, but after about 300ms from the beginning of the motion, the puck position changes suddenly. In response to this, the robot replans the trajectory from the point on the actually performed trajectory located a few tens of milliseconds in the future (to compensate for the nondeterministic communication times and the fact that the used operating system is not real-time) and then waits until the vicinity of this point is reached and switches to the new plan. As it is shown in Fig~\ref{fig:ah_sim_replanning}, the robot smoothly changes between plans and is able to score the goal with the replanned trajectory.

\vspace{0.2cm}
\subsection{Planning high-speed Air Hockey hitting on real robot}

The most important test of the quality of the proposed solution in robotics is the experimental evaluation on a real robot. This is especially important, because of the well-known problem of the \textit{reality gap}, which is common in systems that use machine learning in simulation or learn from a dataset of simulated examples \cite{rss2020reality}. To evaluate if this gap exists in our solution, we used exactly the same neural network for the experiments on the real robot as in the simulation, without any additional learning. 

\begin{figure}[t]
    \centering
    \includegraphics[width=\linewidth]{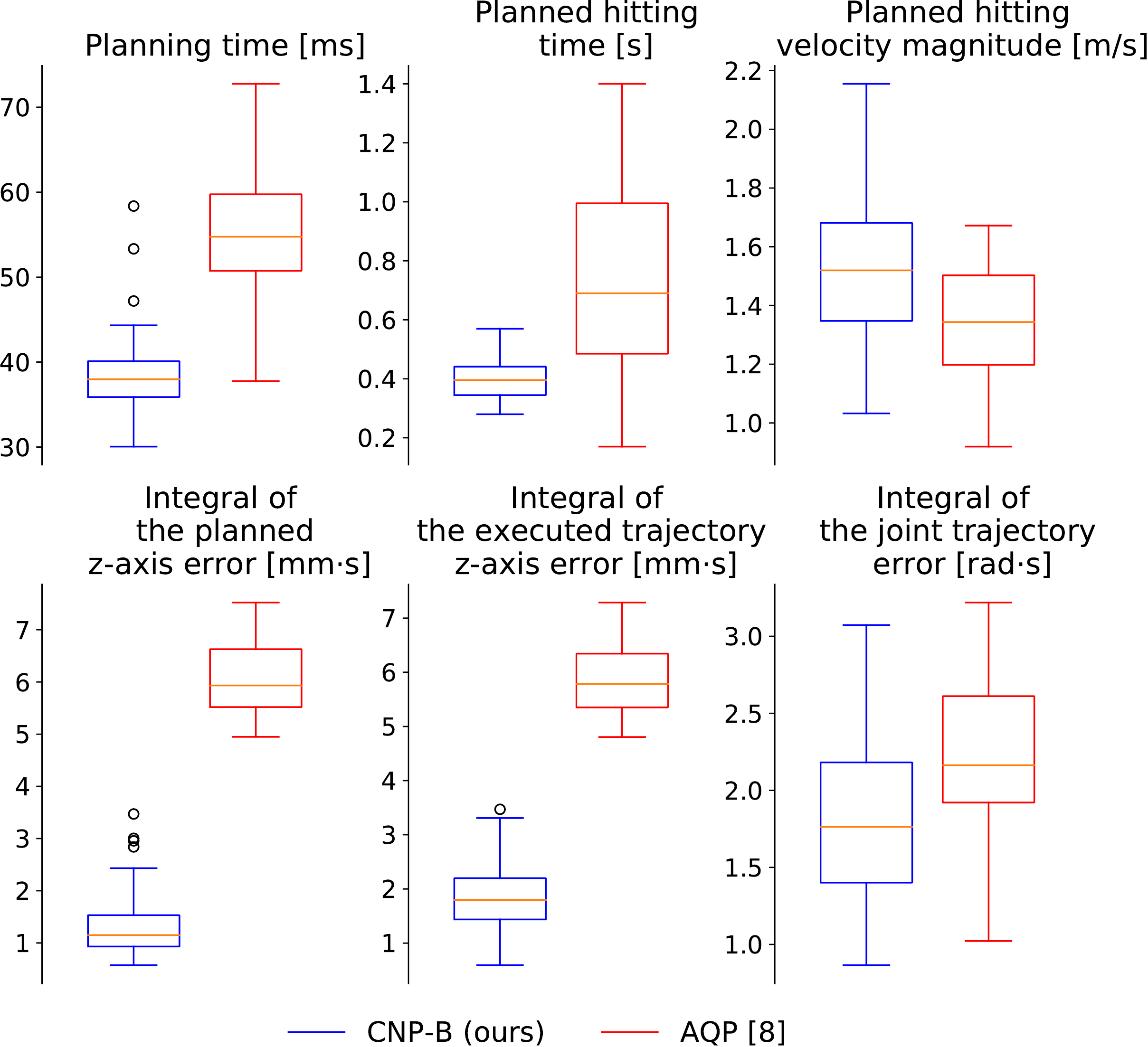}
    \vspace{0.25cm}
    \caption{Planners statistics on the task of hitting in the real robotic Air Hockey.}
    \label{fig:ah_real_stats}
\end{figure}

\vspace{0.5cm}
\subsubsection{Quantitative comparison with state-of-the-art} Similarly, like in simulation, we perform the test of hitting the puck towards the goal, starting from a steady-still manipulator in a base configuration. For each scenario, the puck is placed on 1 of the 110 predefined puck positions (10$\times$11 grid).
Experiments in simulation showed us clearly that \gls{aqp} was the only baseline able to compute safe to follow plans and to compete with the \gls{ours}. Therefore, in experiments on the real robot, we compared our proposed solution only to the \gls{aqp}. Statistical comparison is presented in Fig.~\ref{fig:ah_real_stats}. The mean values of the planning time (time needed to plan both hitting and return movements) and hitting movement time correspond with the one obtained in the simulation. However, in this plot, we can see that \gls{ours} is characterized by a much smaller variance. Moreover, the planned hitting velocity magnitude is higher for the proposed solution, due to the fact that \gls{aqp} method scales down the hitting velocity if it cannot find a feasible plan. The biggest advantage of the proposed planner is visible in terms of the z-axis error, as the generated plans are much closer to the table surface. Also, the trajectory tracking errors are smaller for \gls{ours}, despite faster trajectories.

\begin{figure}[t]
    \centering
    \includegraphics[width=\linewidth]{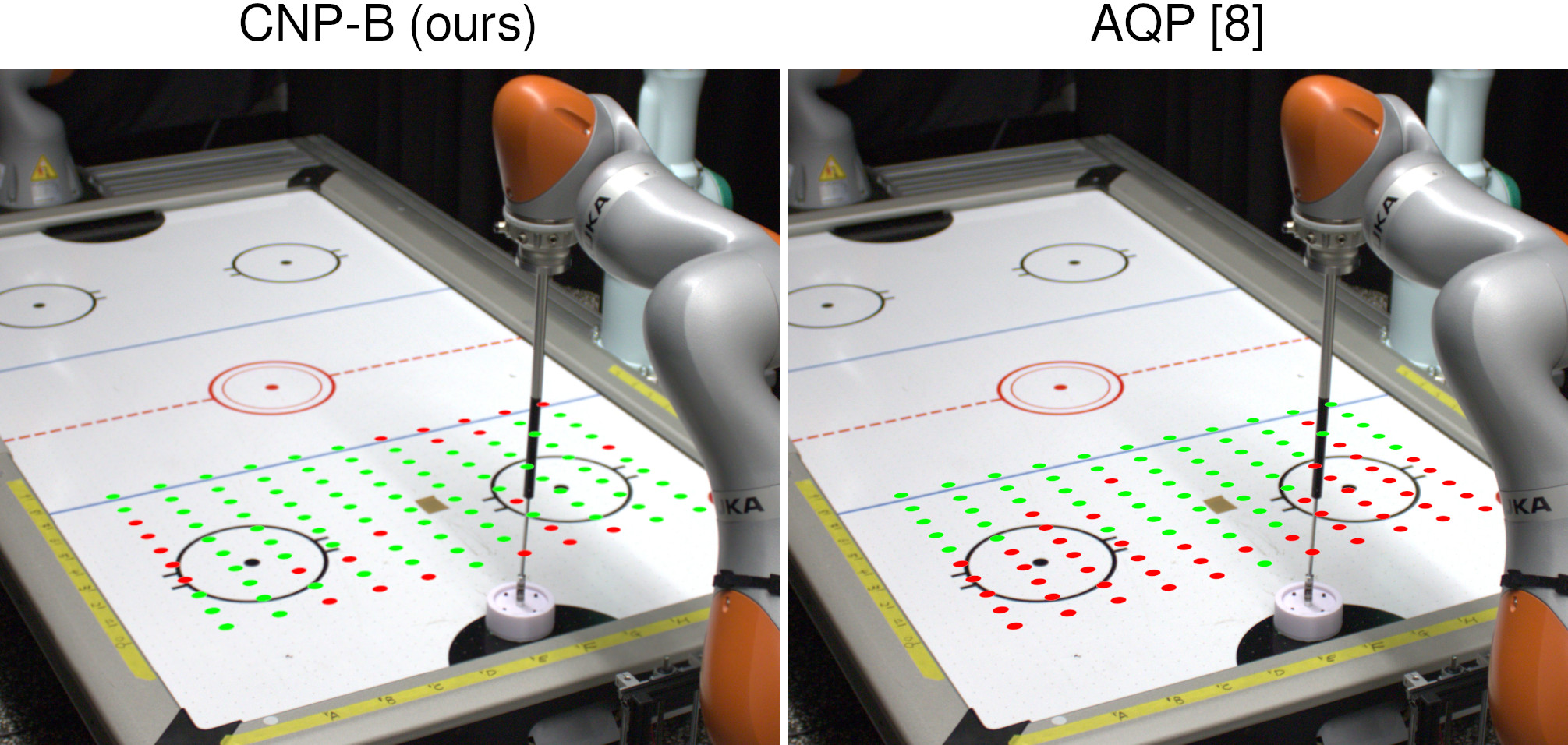}
    \caption{Visualization of the puck positions in the scenarios on which \gls{ours} and \gls{aqp} palnenrs were evaluated. The green dots represent scored goals (success), while the red ones missed shots (failure).}
    \label{fig:hitting_scatter}
\end{figure}

Nevertheless, from the task point of view, the most important metric (besides safety) is the ratio of scored goals to all attempts. In this category, \gls{ours} outperforms \gls{aqp}, by reaching the ratio of 78.2\% compared to 52.7\%. In Fig.~\ref{fig:hitting_scatter} we present the grid of puck positions and indicate the scored goal from this position with green and miss with red. It is visible that \gls{aqp} has problems with executing plans for the puck close to the corners of the table, while \gls{ours} errors seem not to show any particular correlation with the geometry of the playing field. We hypothesize that the few unsuccessful hits by \gls{ours} are related to the mechanical setup of the physical system (particularly the mallet and its attachment to the robot), thus illustrating rather the {\em reality gap} problem.  

\vspace{0.2cm}
\subsubsection{Trick shots} 
\begin{figure*}[tb]
    \centering
    \includegraphics[width=\linewidth]{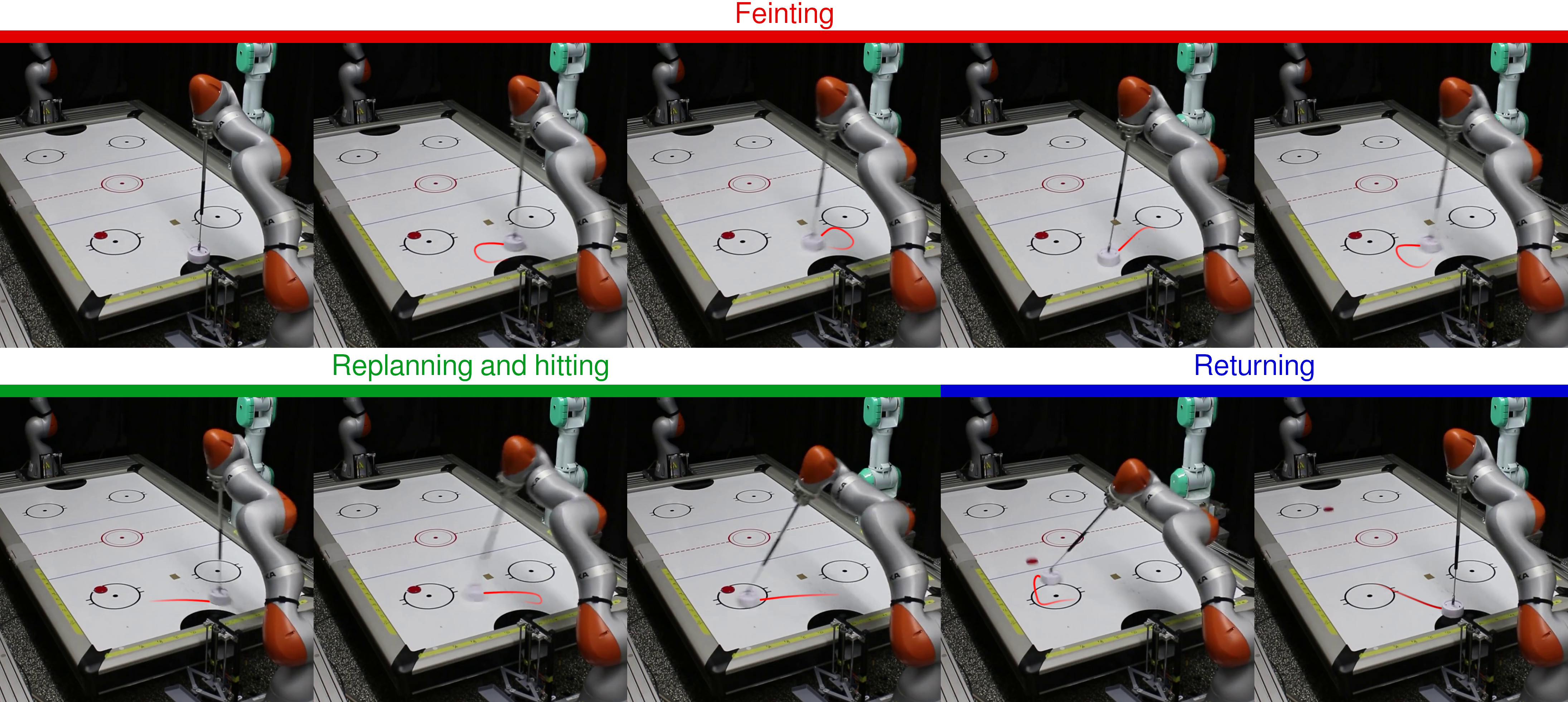}
    \caption{Fast on-the-fly motion replanning can be used to smoothly change the robot's behavior from feinting to striking almost instantaneously.}
    \label{fig:lissajous_hit}
\end{figure*}

The ability of the proposed solution to rapidly plan and replan robot motions, which we have shown in simulation, is also easily transferable to the real robot without any further learning. In Fig.~\ref{fig:lissajous_hit} we present a sequence of frames from the scenario where the robot starts making feinting movements to confuse the opponent, and after some time, computes the new hitting trajectory, starting from non-zero velocity and acceleration, and scores a goal. This kind of dynamic replanning behavior is possible only because our proposed solution plans within a very short and almost constant time, and is able to plan from non-zero boundary conditions imposed on velocity and acceleration.

\vspace{0.2cm}
\subsection{Discussion}
In this section, we want to analyze the strengths and weaknesses of our approach w.r.t. the state-of-the-art planners.
Our experiments have shown that our methodology outperforms every state-of-the-art method under all metrics. 
We believe this performance increase is due to three crucial aspects of our approach: 
(i) the handling of constraints, 
(ii) the use of flexible trajectory parameterization, 
(iii) the learning setup.

First, we treat the constraints satisfaction problem as loss minimization, instead of a hard constraint that should be satisfied at any time. This treatment allows us to accept small violations both during training and deployment. While in some scenarios, small trajectory violations are a major problem, these violations are acceptable for most practical setups. Unfortunately, our handling of the constraints does not guarantee the path planned will satisfy all the constraints. However, it is fairly easy to verify that a planned trajectory complies with the requirements and abort the plan if necessary. State-of-the-art planners handle constraints in various ways, which we described in Section~\ref{sec:related_work}, however, none of them is perfect. The \gls{sst} and \gls{mpcmpnet} algorithms satisfy the robot kinodynamic constraints by relying on the control space sampling. However, for systems with many degrees of freedom, it leads to slow planning. Moreover, sampling-based planners have innate problems when other task-related constraints are present, as they are represented as obstacles that form narrow passages, which results in a further slowdown. In response to this problem, the \gls{cbirrt} algorithm was introduced, ensuring geometrical constraint satisfaction using projection on the constraint manifold. Unfortunately, this planner cannot properly handle the kinodynamic constraints in dynamic motion because it only generates paths.
In turn, optimization-based methods can handle constraints very well, but if antagonistic constraints are present, they may cause the optimization to stuck at a local minimum.

Second, thanks to our B-spline parameterization, we can easily ensure that the plans will always start from the current state and reach the planned one in every trajectory computed by the network. While this property could cause issues in the learning process and plan generation, our B-spline parametrization decouples the geometrical path and the speed of execution by learning a spline for the time coordinate. This decoupling gives us great flexibility and simplifies the learning process. Using parametrized trajectories simplifies the structure of the neural network, avoiding complex learning procedures required for recurrent networks. Also, it is easy to tune the number of control points of the spline to control the maximum allowed complexity of the path.

Finally, our learning setup is simple, as it does not require any expert dataset besides the models of the robot and environment. Furthermore, it is easy to generalize to different tasks and settings. In principle, if we provide a sufficiently large training set, this approach could process any relevant information to solve the task, allowing for advanced obstacle avoidance and maneuvering e.g., as done in~\cite{kicki2022bspline}. However, there is no practical way to guarantee that the generated plans will not violate any of the constraints. 
This lack of guarantees can be potentially seen as a drawback w.r.t. other methods, as they can adapt on the fly and optimize the trajectory, ensuring that the solution found is feasible. However, it is always possible to improve the trajectory generated by our method with further optimization e.g., using the TrajOpt framework. Further optimization of the generated trajectory is feasible because the planning time of the proposed method is negligible compared to basic trajectory optimization. 
Another important gain from the short planning time, combined with a constant amount of computation and boundary conditions satisfaction thanks to B-spline trajectory representation, is the ability to feasible and smooth replanning on the fly. To the best of our knowledge, it is the first time this kind of dynamic smooth replanning is presented for problems of this complexity, along with a demonstration on a real robot.

\section{Conclusions}
In this paper, we presented a complete learning solution to kinodynamic planning under arbitrary constraints. Exploiting Deep Neural Networks, we are able to learn a planning function generating plans lying in close vicinity of the constraint manifold. Our approach is easy to implement, computes plans with minimal computational requirements, and can replan from arbitrary configurations. Our method is flexible and can adapt to many different tasks, and it does not require any reference trajectories for training. While we cannot ensure the safety of all the plans generated by the neural network, it is always possible to perform post-processing steps and discard the plans that do not satisfy the constraints.
Our real robot experiment shows that the proposed method can be successfully used in real-world tasks, outperforming existing ad-hoc solutions. 
\gls{ours} can be seen as a solution bridging the gap between a reactive neural controller and a classic motion planning algorithm. 
It is fast as the reactive neural controller but computes the whole plan at once, like a planning method, which is important from the safety point of view.

In future work, we would like to include some representation of the constraint manifold in the input of the neural network to be able to plan on multiple manifolds without retraining. 
It would also be interesting to check whether \gls{ours} can be applied as a motion generator, plugged into a higher-level reinforcement learning setup as a parametrizable action, e.g. to learn how to play the whole game of Air Hockey.

\section*{Acknowledgments}
\noindent This project was supported by the CSTT fund from Huawei Tech R\&D (UK). Part of the research presented has been founded by China Scholarship Council (No. 201908080039).
Research presented in this paper has been supported by the German Federal Ministry of Education and Research (BMBF) within a subproject ``Modeling and exploration of the operational area, design of the AI assistance as well as legal aspects of the use of technology'' of the collaborative KIARA project (grant no. 13N16274).
The work of P. Kicki has been supported by the Polish National Agency for Academic Exchange (NAWA) under the STER programme ``Towards Internationalization of Poznan University of Technology Doctoral School'' (2022-2024). P. Skrzypczy{\'n}ski is supported by TAILOR, a project funded by EU Horizon 2020 under GA No. 952215.

\bibliographystyle{IEEEtran}
\bibliography{bibliography}

\begin{thebibliography}{10}
\providecommand{\url}[1]{#1}
\csname url@rmstyle\endcsname
\providecommand{\newblock}{\relax}
\providecommand{\bibinfo}[2]{#2}
\providecommand\BIBentrySTDinterwordspacing{\spaceskip=0pt\relax}
\providecommand\BIBentryALTinterwordstretchfactor{4}
\providecommand\BIBentryALTinterwordspacing{\spaceskip=\fontdimen2\font plus
\BIBentryALTinterwordstretchfactor\fontdimen3\font minus
  \fontdimen4\font\relax}
\providecommand\BIBforeignlanguage[2]{{%
\expandafter\ifx\csname l@#1\endcsname\relax
\typeout{** WARNING: IEEEtran.bst: No hyphenation pattern has been}%
\typeout{** loaded for the language `#1'. Using the pattern for}%
\typeout{** the default language instead.}%
\else
\language=\csname l@#1\endcsname
\fi
#2}}

\bibitem{kawato1994teaching}
M.~Kawato, F.~Gandolfo, H.~Gomi, and Y.~Wada, ``Teaching by showing in kendama
  based on optimization principle,'' in \emph{International Conference on
  Artificial Neural Networks}.\hskip 1em plus 0.5em minus 0.4em\relax Springer,
  1994, pp. 601--606.

\bibitem{kober2008policy}
J.~Kober and J.~Peters, ``Policy search for motor primitives in robotics,''
  \emph{Advances in neural information processing systems}, vol.~21, 2008.

\bibitem{mulling2011biomimetic}
K.~M{\"u}lling, J.~Kober, and J.~Peters, ``A biomimetic approach to robot table
  tennis,'' \emph{Adaptive Behavior}, vol.~19, no.~5, pp. 359--376, 2011.

\bibitem{buchler2022learning}
D.~B{\"u}chler, S.~Guist, R.~Calandra, V.~Berenz, B.~Sch{\"o}lkopf, and
  J.~Peters, ``Learning to play table tennis from scratch using muscular
  robots,'' \emph{IEEE Transactions on Robotics}, 2022.

\bibitem{ploeger2021high}
K.~Ploeger, M.~Lutter, and J.~Peters, ``High acceleration reinforcement
  learning for real-world juggling with binary rewards,'' in \emph{Conference
  on Robot Learning}.\hskip 1em plus 0.5em minus 0.4em\relax PMLR, 2021, pp.
  642--653.

\bibitem{von2021analytical}
F.~von Drigalski, D.~Joshi, T.~Murooka, K.~Tanaka, M.~Hamaya, and Y.~Ijiri,
  ``An analytical diabolo model for robotic learning and control,'' in
  \emph{2021 IEEE International Conference on Robotics and Automation
  (ICRA)}.\hskip 1em plus 0.5em minus 0.4em\relax IEEE, 2021, pp. 4055--4061.

\bibitem{namiki2013hierarchical}
A.~Namiki, S.~Matsushita, T.~Ozeki, and K.~Nonami, ``Hierarchical processing
  architecture for an air-hockey robot system,'' in \emph{2013 IEEE
  International Conference on Robotics and Automation}.\hskip 1em plus 0.5em
  minus 0.4em\relax IEEE, 2013, pp. 1187--1192.

\bibitem{liu2021efficient}
P.~Liu, D.~Tateo, H.~Bou-Ammar, and J.~Peters, ``Efficient and reactive
  planning for high speed robot air hockey,'' in \emph{2021 IEEE/RSJ
  International Conference on Intelligent Robots and Systems (IROS)}.\hskip 1em
  plus 0.5em minus 0.4em\relax IEEE, 2021, pp. 586--593.

\bibitem{kingston2018sampling}
Z.~Kingston, M.~Moll, and L.~E. Kavraki, ``Sampling-based methods for motion
  planning with constraints,'' \emph{Annual Review of Control, Robotics, and
  Autonomous Systems}, vol.~1, no.~1, pp. 159--185, 2018.

\bibitem{kingston2019exploring}
Z.~Kingston, M.~Moll, and L.~E.~Kavraki, ``Exploring implicit spaces for
  constrained sampling-based planning,'' \emph{The International Journal of
  Robotics Research}, vol.~38, no. 10-11, pp. 1151--1178, 2019.

\bibitem{berenson2009manipulation}
D.~Berenson, S.~S. Srinivasa, D.~Ferguson, and J.~J. Kuffner, ``Manipulation
  planning on constraint manifolds,'' in \emph{2009 IEEE International
  Conference on Robotics and Automation}, 2009, pp. 625--632.

\bibitem{Bordalba2021atlasKinodynamicRRT}
R.~Bordalba, L.~Ros, and J.~M. Porta, ``A randomized kinodynamic planner for
  closed-chain robotic systems,'' \emph{IEEE Transactions on Robotics},
  vol.~37, no.~1, pp. 99--115, 2021.

\bibitem{Xie2020kinodymanicFactorGraphs}
\BIBentryALTinterwordspacing
M.~Xie and F.~Dellaert, ``Batch and incremental kinodynamic motion planning
  using dynamic factor graphs,'' 2020. [Online]. Available:
  \url{https://arxiv.org/abs/2005.12514}
\BIBentrySTDinterwordspacing

\bibitem{liu2021robot}
\BIBentryALTinterwordspacing
P.~Liu, D.~Tateo, H.~B. Ammar, and J.~Peters, ``Robot reinforcement learning on
  the constraint manifold,'' in \emph{5th Annual Conference on Robot Learning},
  2021. [Online]. Available: \url{https://openreview.net/forum?id=zwo1-MdMl1P}
\BIBentrySTDinterwordspacing

\bibitem{kicki2022bspline}
P.~Kicki and P.~Skrzypczyński, ``Speeding up deep neural network-based
  planning of local car maneuvers via efficient b-spline path construction,''
  in \emph{International Conference on Robotics and Automation (ICRA)}, 2022,
  pp. 4422--4428.

\bibitem{kicki2021learning}
P.~Kicki, T.~Gawron, K.~{\'C}wian, M.~Ozay, and P.~Skrzypczy{\'n}ski,
  ``Learning from experience for rapid generation of local car maneuvers,''
  \emph{Engineering Applications of Artificial Intelligence}, vol. 105, p.
  104399, 2021.

\bibitem{prm}
L.~Kavraki, P.~Svestka, J.-C. Latombe, and M.~Overmars, ``Probabilistic
  roadmaps for path planning in high-dimensional configuration spaces,''
  \emph{IEEE Transactions on Robotics and Automation}, vol.~12, no.~4, pp.
  566--580, 1996.

\bibitem{rrt}
S.~M. LaValle, ``Rapidly-exploring random trees : a new tool for path
  planning,'' \emph{The annual research report}, 1998.

\bibitem{rrtstar}
S.~Karaman and E.~Frazzoli, ``Sampling-based algorithms for optimal motion
  planning,'' \emph{The International Journal of Robotics Research}, vol.~30,
  no.~7, pp. 846--894, 2011.

\bibitem{bfmt}
J.~A. Starek, J.~V. Gomez, E.~Schmerling, L.~Janson, L.~Moreno, and M.~Pavone,
  ``An asymptotically-optimal sampling-based algorithm for bi-directional
  motion planning,'' in \emph{2015 IEEE/RSJ International Conference on
  Intelligent Robots and Systems (IROS)}, 2015, pp. 2072--2078.

\bibitem{bitstar_conf}
J.~D. {Gammell}, S.~S. {Srinivasa}, and T.~D. {Barfoot}, ``Batch informed trees
  {(BIT$^*$)}: Sampling-based optimal planning via the heuristically guided
  search of implicit random geometric graphs,'' in \emph{IEEE International
  Conference on Robotics and Automation}, 2015, pp. 3067--3074.

\bibitem{NonholonomicRRT}
S.~{Karaman} and E.~{Frazzoli}, ``Sampling-based optimal motion planning for
  non-holonomic dynamical systems,'' in \emph{{IEEE} International Conference
  on Robotics and Automation}, Karlsruhe, 2013, pp. 5041--5047.

\bibitem{bitstar}
J.~D. Gammell, T.~D. Barfoot, and S.~S. Srinivasa, ``Batch informed trees
  ({BIT$^*$}): Informed asymptotically optimal anytime search,'' \emph{The
  International Journal of Robotics Research}, vol.~39, no.~5, pp. 543--567,
  2020.

\bibitem{chomp}
M.~Zucker, N.~Ratliff, A.~D. Dragan, M.~Pivtoraiko, M.~Klingensmith, C.~M.
  Dellin, J.~A. Bagnell, and S.~S. Srinivasa, ``{CHOMP}: Covariant hamiltonian
  optimization for motion planning,'' \emph{The International Journal of
  Robotics Research}, vol.~32, no. 9-10, pp. 1164--1193, 2013.

\bibitem{trajOpt}
J.~Schulman, J.~Ho, A.~X. Lee, I.~Awwal, H.~Bradlow, and P.~Abbeel, ``Finding
  locally optimal, collision-free trajectories with sequential convex
  optimization,'' \emph{Robotics: Science and Systems IX}, 2013.

\bibitem{mukadam2018gpmp2}
\BIBentryALTinterwordspacing
M.~Mukadam, J.~Dong, X.~Yan, F.~Dellaert, and B.~Boots, ``Continuous-time
  gaussian process motion planning via probabilistic inference,'' \emph{The
  International Journal of Robotics Research}, vol.~37, no.~11, p. 1319–1340,
  Sep 2018. [Online]. Available:
  \url{http://dx.doi.org/10.1177/0278364918790369}
\BIBentrySTDinterwordspacing

\bibitem{constrainedSQP}
R.~Bonalli, A.~Cauligi, A.~Bylard, T.~Lew, and M.~Pavone, ``Trajectory
  optimization on manifolds: A theoretically-guaranteed embedded sequential
  convex programming approach,'' in \emph{Proceedings of Robotics: Science and
  Systems}, Freiburg, Germany, June 2019.

\bibitem{constrainedCHOMP}
A.~D. Dragan, N.~D. Ratliff, and S.~S. Srinivasa, ``Manipulation planning with
  goal sets using constrained trajectory optimization,'' in \emph{2011 IEEE
  International Conference on Robotics and Automation}, 2011, pp. 4582--4588.

\bibitem{Howell2019ALTRO}
T.~A. Howell, B.~E. Jackson, and Z.~Manchester, ``{ALTRO}: A fast solver for
  constrained trajectory optimization,'' in \emph{2019 IEEE/RSJ International
  Conference on Intelligent Robots and Systems (IROS)}, 2019, pp. 7674--7679.

\bibitem{Li2004iLQR}
W.~Li and E.~Todorov, ``Iterative linear quadratic regulator design for
  nonlinear biological movement systems.'' in \emph{Proceedings of the 1st
  International Conference on Informatics in Control, Automation and Robotics,
  (ICINCO 2004)}, vol.~1, 01 2004, pp. 222--229.

\bibitem{constrainedSBMPrelaxation1}
M.~Bonilla, E.~Farnioli, L.~Pallottino, and A.~Bicchi, ``Sample-based motion
  planning for robot manipulators with closed kinematic chains,''
  \emph{Proceedings - IEEE International Conference on Robotics and
  Automation}, vol. 2015, pp. 2522--2527, 06 2015.

\bibitem{constrainedSBMPrelaxation2}
M.~Bonilla, L.~Pallottino, and A.~Bicchi, ``Noninteracting constrained motion
  planning and control for robot manipulators,'' in \emph{2017 IEEE
  International Conference on Robotics and Automation (ICRA)}, 2017, pp.
  4038--4043.

\bibitem{szkandera2020}
J.~Szkandera, I.~Kolingerov{\'a}, and M.~Ma{\v{n}}{\'a}k, ``Narrow passage
  problem solution for motion planning,'' in \emph{Computational Science --
  ICCS 2020}, V.~V. Krzhizhanovskaya, G.~Z{\'a}vodszky, M.~H. Lees, J.~J.
  Dongarra, P.~M.~A. Sloot, S.~Brissos, and J.~Teixeira, Eds.\hskip 1em plus
  0.5em minus 0.4em\relax Cham: Springer International Publishing, 2020, pp.
  459--470.

\bibitem{wang2015constrained}
J.~Wang, J.~Lee, and J.~Kim, ``Constrained motion planning for robot
  manipulators using local geometric information,'' \emph{Advanced Robotics},
  vol.~29, no.~24, pp. 1611--1623, 2015.

\bibitem{TBRRT}
B.~Kim, T.~Um, C.~Suh, and F.~Park, ``Tangent bundle {RRT}: A randomized
  algorithm for constrained motion planning,'' \emph{Robotica}, vol.~34, pp.
  202--225, 01 2016.

\bibitem{atlasRRT}
L.~Jaillet and J.~M. Porta, ``Path planning under kinematic constraints by
  rapidly exploring manifolds,'' \emph{IEEE Transactions on Robotics}, vol.~29,
  no.~1, pp. 105--117, 2013.

\bibitem{Perez2012LQRRRT}
A.~Perez, R.~Platt, G.~Konidaris, L.~Kaelbling, and T.~Lozano-Perez,
  ``{LQR-RRT$^*$}: Optimal sampling-based motion planning with automatically
  derived extension heuristics,'' in \emph{2012 IEEE International Conference
  on Robotics and Automation}, 2012, pp. 2537--2542.

\bibitem{Stoneman2014NLPRRT}
S.~Stoneman and R.~Lampariello, ``Embedding nonlinear optimization in {RRT} for
  optimal kinodynamic planning,'' in \emph{53rd IEEE Conference on Decision and
  Control}, 2014, pp. 3737--3744.

\bibitem{Primatesta2021MPCRRT}
\BIBentryALTinterwordspacing
S.~Primatesta, A.~Osman, and A.~Rizzo, ``{MP-RRT\#}: a model predictive
  sampling-based motion planning algorithm for unmanned aircraft systems,''
  \emph{Journal of Intelligent {\&} Robotic Systems}, vol. 103, no.~4, p.~59,
  Nov 2021. [Online]. Available:
  \url{https://doi.org/10.1007/s10846-021-01501-3}
\BIBentrySTDinterwordspacing

\bibitem{Zheng2021KinoRRT}
D.~Zheng and P.~Tsiotras, ``Accelerating kinodynamic {RRT} through
  dimensionality reduction,'' in \emph{2021 IEEE/RSJ International Conference
  on Intelligent Robots and Systems (IROS)}, 2021, pp. 3674--3680.

\bibitem{sst}
\BIBentryALTinterwordspacing
Y.~Li, Z.~Littlefield, and K.~E. Bekris, ``Asymptotically optimal
  sampling-based kinodynamic planning,'' \emph{The International Journal of
  Robotics Research}, vol.~35, no.~5, pp. 528--564, 2016. [Online]. Available:
  \url{https://doi.org/10.1177/0278364915614386}
\BIBentrySTDinterwordspacing

\bibitem{Cefalo2014TCMP}
M.~Cefalo and G.~Oriolo, ``Dynamically feasible task-constrained motion
  planning with moving obstacles,'' in \emph{2014 IEEE International Conference
  on Robotics and Automation (ICRA)}, 2014, pp. 2045--2050.

\bibitem{Berenson2012remember}
D.~Berenson, P.~Abbeel, and K.~Goldberg, ``A robot path planning framework that
  learns from experience,'' in \emph{2012 IEEE International Conference on
  Robotics and Automation}, 2012, pp. 3671--3678.

\bibitem{Zhang2018distribution}
C.~Zhang, J.~Huh, and D.~D. Lee, ``Learning implicit sampling distributions for
  motion planning,'' in \emph{2018 IEEE/RSJ International Conference on
  Intelligent Robots and Systems (IROS)}, 2018, pp. 3654--3661.

\bibitem{Huh2018rrtq}
J.~Huh and D.~D. Lee, ``Efficient sampling with q-learning to guide rapidly
  exploring random trees,'' \emph{IEEE Robotics and Automation Letters},
  vol.~3, no.~4, pp. 3868--3875, 2018.

\bibitem{Molina2020learnAndLink}
D.~Molina, K.~Kumar, and S.~Srivastava, ``Learn and link: Learning critical
  regions for efficient planning,'' in \emph{2020 IEEE International Conference
  on Robotics and Automation (ICRA)}, 2020, pp. 10\,605--10\,611.

\bibitem{Cheng2020distribution}
R.~Cheng, K.~Shankar, and J.~W. Burdick, ``Learning an optimal sampling
  distribution for efficient motion planning,'' in \emph{2020 IEEE/RSJ
  International Conference on Intelligent Robots and Systems (IROS)}, 2020, pp.
  7485--7492.

\bibitem{Quershi2021MPNet}
A.~H. Qureshi, Y.~Miao, A.~Simeonov, and M.~C. Yip, ``Motion planning networks:
  Bridging the gap between learning-based and classical motion planners,''
  \emph{IEEE Transactions on Robotics}, vol.~37, no.~1, pp. 48--66, 2021.

\bibitem{Lembono2021constrainedDistribution}
T.~S. Lembono, E.~Pignat, J.~Jankowski, and S.~Calinon, ``Learning constrained
  distributions of robot configurations with generative adversarial network,''
  \emph{IEEE Robotics and Automation Letters}, vol.~6, no.~2, pp. 4233--4240,
  2021.

\bibitem{CoMPNetX}
A.~H. Qureshi, J.~Dong, A.~Baig, and M.~C. Yip, ``Constrained motion planning
  networks x,'' \emph{IEEE Transactions on Robotics}, pp. 1--19, 2021.

\bibitem{Wolfslag2018KMPLearnSteering}
W.~J. Wolfslag, M.~Bharatheesha, T.~M. Moerland, and M.~Wisse, ``{RRT-CoLearn}:
  Towards kinodynamic planning without numerical trajectory optimization,''
  \emph{IEEE Robotics and Automation Letters}, vol.~3, no.~3, pp. 1655--1662,
  2018.

\bibitem{Atreya2022KMPLearnSteering}
P.~Atreya and J.~Biswa, ``State supervised steering function for sampling-based
  kinodynamic planning,'' in \emph{2022 International Conference on Autonomous
  Agents and Multiagent Systems (AAMAS)}, 2022, pp. 35--43.

\bibitem{Yavari2019KMPLazy}
M.~Yavari, K.~Gupta, and M.~Mehrandezh, ``Lazy steering {RRT}: An optimal
  constrained kinodynamic neural network based planner with no in-exploration
  steering,'' in \emph{2019 19th International Conference on Advanced Robotics
  (ICAR)}, 2019, pp. 400--407.

\bibitem{li2021mpcmpnet}
L.~Li, Y.~Miao, A.~H. Qureshi, and M.~C. Yip, ``{MPC-MPNet}: Model-predictive
  motion planning networks for fast, near-optimal planning under kinodynamic
  constraints,'' \emph{IEEE Robotics and Automation Letters}, vol.~6, no.~3,
  pp. 4496--4503, 2021.

\bibitem{bertsekas_largrange}
D.~Bertsekas, \emph{Nonlinear Programming}.\hskip 1em plus 0.5em minus
  0.4em\relax Athena Scientific, 1999.

\bibitem{stooke2020responsive}
A.~Stooke, J.~Achiam, and P.~Abbeel, ``Responsive safety in reinforcement
  learning by pid lagrangian methods,'' in \emph{International Conference on
  Machine Learning}.\hskip 1em plus 0.5em minus 0.4em\relax PMLR, 2020, pp.
  9133--9143.

\bibitem{carpentier2019pinocchio}
J.~Carpentier, G.~Saurel, G.~Buondonno, J.~Mirabel, F.~Lamiraux, O.~Stasse, and
  N.~Mansard, ``The {Pinocchio} c++ library -- a fast and flexible
  implementation of rigid body dynamics algorithms and their analytical
  derivatives,'' in \emph{IEEE International Symposium on System Integrations
  (SII)}, 2019.

\bibitem{struz2017iiwaparams}
\BIBentryALTinterwordspacing
Y.~R. Stürz, L.~M. Affolter, and R.~S. Smith, ``Parameter identification of
  the {KUKA LBR} iiwa robot including constraints on physical feasibility,''
  \emph{IFAC-PapersOnLine}, vol.~50, no.~1, pp. 6863--6868, 2017, 20th IFAC
  World Congress. [Online]. Available:
  \url{https://www.sciencedirect.com/science/article/pii/S2405896317317147}
\BIBentrySTDinterwordspacing

\bibitem{rss2020reality}
S.~Höfer, K.~Bekris, A.~Handa, J.~Higuera, F.~Golemo, M.~Mozifian, C.~Atkeson,
  D.~Fox, K.~Goldberg, J.~Leonard, C.~Liu, J.~Peters, S.~Song, P.~Welinder, and
  M.~White, ``Perspectives on sim2real transfer for robotics: A summary of the
  r:ss 2020 workshop,'' 12 2020.

\bibitem{sucan2012ompl}
I.~A. {\c{S}}ucan, M.~Moll, and L.~E. Kavraki, ``The {O}pen {M}otion {P}lanning
  {L}ibrary,'' \emph{{IEEE} Robotics \& Automation Magazine}, vol.~19, no.~4,
  pp. 72--82, December 2012, \url{https://ompl.kavrakilab.org}.

\end{thebibliography}

\appendix

\subsection{Satisfying B-spline boundary constraints}
\label{ap:bspline}
To satisfy the boundary constraints of the problem using our B-spline trajectory, we define the vector of knots $\knots$ by

\vspace{0.1cm}
{\small
\begin{equation*}
    \knots = \begin{bmatrix} 
    \underbrace{\vphantom{\frac{1}{\numControlPoints-D}}
    0 \quad \ldots \quad 0}_{D+1\, \text{times}} & 
    \underbrace{\frac{1}{\numControlPoints-D} \quad \ldots \quad \frac{\numControlPoints - D - 1}{\numControlPoints-D}}_{\numControlPoints-D-1\, \text{elements}} & \underbrace{\vphantom{\frac{1}{\numControlPoints-D}}
    1 \quad \ldots \quad 1}_{D+1\, \text{times}}\\
    \end{bmatrix},
\end{equation*}
}
\vspace{0.1cm}

where $D$ is the degree of the B-spline and $\numControlPoints$ is the number of the B-spline control points. This design of the knots vector ensures that the resultant B-spline value on the boundaries will be defined directly by the first and last control points. Thus, we can introduce formulas to calculate the first three and last two control points $\qControlPoints$ based on the $\rates$ control points and reaching task constraints

\begin{align}
\label{eq:controlPoints}
    \qControlPoints_0\quad &= \q_0,\nonumber\\
    \qControlPoints_1\quad &= \qControlPoints_0 + \frac{\dq_0}{\tControlPoints_0 \eta_{\Path}},\nonumber\\
    K \quad &= \frac{\ddq_0 - \eta_{\Path} \eta_{\rate} (\qControlPoints_1 - \qControlPoints_0) (\tControlPoints_1 - \tControlPoints_0)\tControlPoints_0}{\pare{\tControlPoints_0}^2}, \nonumber\\
    \qControlPoints_2\quad &= \frac{K - 2\beta_{\Path} \qControlPoints_0 + 3\beta_{\Path} \qControlPoints_1}{\beta_{\Path}},\nonumber\\
    \qControlPoints_{\numQControlPoints - 1} &= \q_d,\nonumber\\
    \qControlPoints_{\numQControlPoints - 2} &= \qControlPoints_{\numQControlPoints - 1} - \frac{\dq_d}{\tControlPoints_{\numTControlPoints - 1} \eta_{\Path}},
\end{align}
\vspace{0.1cm}

where $\eta_{\Path}$, $\beta_{\Path}$ and $\eta_{\rate}$ are defined by
\begin{align*}
    \eta_{\Path} &= \qDegree (\numQControlPoints - \qDegree)^2,\\
    \beta_{\Path} &= \frac{\qDegree(\qDegree - 1)}{2} (\numQControlPoints - \qDegree)^2,\\
    \eta_{\rate} &= \tDegree (\numTControlPoints - \tDegree)^2,
\end{align*}

where $\qDegree$ and $\tDegree$ are the degrees of the $\Paths$ and $\rates$ B-splines respectively, and $\numQControlPoints$ and $\numTControlPoints$ are the numbers of their control points.

\subsection{Experimental evaluation details}

\subsubsection{Heavy object manipulation} 
\label{ap:kinodynamic_task}
~\\
\paragraph{Data preparation} The data generation process for this experiment was done in the following way:
\begin{enumerate}[label=\roman*.]
    \item draw random initial and desired position of the heavy object,
    \item assume that the pedestal boxes have fixed dimensions and are located just beneath the objects,
    \item draw initial guess configuration of the robot,
    \item starting from this configuration, optimize the robot's initial configuration, such that its end-effector position matches the initial position of the heavy object, and the orientation is vertical,
    \item validate if the robot in the initial configuration does not collide with the environment and if it does not violate the torque constraints,
    \item starting from the initial configuration, optimize the robot desired configuration, such that its end-effector position matches the desired position of the heavy object,
    \item validate if the robot in the desired configuration does not collide with the environment and if it does not violate the torque constraints.
\end{enumerate}
The specific parameters values and ranges are shown in Table~\ref{tab:kino_dataset_ranges}, where $o_{z0}, o_{zd}$ represents the object's initial and desired position along $z$-axis, whereas $o_h = \SI{0.15}{\metre}$ is the fixed height of the object.

\paragraph{Additional loss functions}
As we already mentioned in the main text, in this task, to meet the constraints imposed on this task, we introduced 3 additional loss terms i.e. vertical
orientation loss defined by
\begin{equation*}
    \stepManifoldLoss^{O}(s) = \huber(1 - \Rot_{2,2}(\q(s))),
\end{equation*}
where $\Rot_{2,2}$ is the element of the end-effector rotation matrix with an index of $(2, 2)$,
robot collision loss defined by
\begin{equation*}
    \stepManifoldLoss^{E_r}(s) = \huber\left(\sum_{p \in \FK_{kc}(\q(s))}\relu(0.15 - d(p, E))\right),
\end{equation*}
where $E$ represents the set of the collision objects in the environment (pedestals), $\FK_{kc}$ is a set of points in the workspace located along the kinematic chain (representation of the robot geometry), and $d(X, Y)$ is a Euclidean distance between $X$ and $Y$, 
and finally, object collision loss
\begin{equation*}
    \stepManifoldLoss^{E_o}(s) = \huber\left(\sum_{p \in \FK_{o}(\q(s))}  \relu(d(p, E) \cdot \ind(p, E))\right),
\end{equation*}
where $\FK_{o}$ represents the set of points that belong to the handled object and $\ind(X, Y)$ is an indicator function, which is equal to 1 if $X \in Y$ and 0 otherwise.
In our experiments, we defined environment $E$ as two cuboids defined in Tab.~\ref{tab:kino_dataset_ranges}. The heavy object handled by the robot is a cuboid with dimensions $0.2 \times 0.2 \times 0.3$ m, which for collision-checking purposes is represented by its corners. The robot itself is represented by the positions of the joints in the workspace and points linearly interpolated between them, such that no point lies further than \SI{10}{\cm} from its neighbors.

\begin{table}[!bt]
\renewcommand{\arraystretch}{2.5}
\centering
\caption{Parameters of the data generation procedure for heavy object manipulation task.}
\begin{tabular}{c|c}
\hline
Parameter & Value\\
\hline
\makecell{Initial \\ object position} & $(x,y,z) \in [0.2; 0.6] \times [-0.6; -0.3] \times [0.2; 0.5]$\\
\makecell{Desired \\ object position} & $(x,y,z) \in [0.2; 0.6] \times [0.3; 0.6] \times [0.2; 0.5]$\\
Pedestal 1 & $\begin{aligned}
    \{(x, y, z)\,|\,& 0.2 \leq x \leq 0.6, \\
    & -0.6 \leq y \leq -0.3, z \leq o_{z0} - o_h\}
\end{aligned}$\\[0.3cm]
Pedestal 2 & $\begin{aligned}
    \{(x, y, z)\,|\,& 0.2 \leq x \leq 0.6, \\
    & 0.3 \leq y \leq 0.6, z \leq o_{zd} - o_h\}
\end{aligned}$\\[0.3cm]
\makecell{Initial guess\\ robot configuration} & $\begin{aligned} q \in\, & [-\pitwo; \pitwo] \times [0; \pitwo] \times [-\pitwo; \pitwo] \\ & \times [\pitwo; 0] \times [-\pitwo; \pitwo]^2 \times [-\pi;\pi] \end{aligned}$ \\[0.5cm]
\hline
\end{tabular}
\label{tab:kino_dataset_ranges}
\end{table}

\subsubsection{Simulated Air Hockey hitting}
\label{ap:ah_sim_task}
\paragraph{Data preparation} The data generation process for this experiment was done in the following way:
\begin{enumerate}[label=\roman*.]
    \item draw random initial and desired position of the mallet, such that they are at least \SI{10}{\cm} apart,
    \item starting from the base configuration, optimize the robot's initial configuration, such that its end-effector position matches the initial position of the mallet,
    \item using the desired mallet position and goal position, define the desired hitting angle, and add noise to it,
    \item compute the desired joint velocity of the robot that maximizes the manipulability along the hitting direction~\cite{liu2021efficient},
    \item in half of the cases randomly scale the magnitude of the desired joint velocity, and set it to maximal in the other half,
    \item validate the possibility of performing the hit, by analyzing if it is possible to avoid a collision after the hit, i.e. if the point defined by
    $p_h = p_d + v_h \cdot \SI{50}{\ms}$
    lies on the table surface, where $p_d$ is desired hitting point, and $v_h$ is the hitting velocity in the workspace.
\end{enumerate}
The specific parameter values and ranges are shown in Table~\ref{tab:ah_dataset_ranges}.

\begin{table}[!bt]
\renewcommand{\arraystretch}{2}
\centering
\caption{Parameters of the data generation procedure for Air Hockey hitting task.}
\begin{tabular}{c|c}
\hline
Parameter & Value\\
\hline
\makecell{Initial \\ mallet position} & $(x,y,z) \in [0.6; 0.7] \times [-0.05; 0.05] \times [0.155; 0.165]$\\
\makecell{Desired \\ mallet position} & $(x,y,z) \in [0.65; 1.3] \times [-0.45; 0.45] \times \{0.16\}$\\
\makecell{Base robot \\ configuration} & $q_0 = [ 0 \quad 0.697 \quad 0 \quad -0.505 \quad 0 \quad 1.93 ]$\\
\hline
\end{tabular}
\label{tab:ah_dataset_ranges}
\end{table}

\paragraph{Additional loss functions}
As we already mentioned in the main text, in this task, to meet the constraints imposed on this task, we introduced an additional constraint manifold loss term, which is responsible to maintain the mallet position on the table surface. We define this loss function as the sum of the losses in $x, y, z$ directions
\begin{equation*}
    \stepManifoldLoss^{\mathcal{T}}(s) = \stepManifoldLoss^{\mathcal{T}_x}(s) + \stepManifoldLoss^{\mathcal{T}_y}(s)+ \stepManifoldLoss^{\mathcal{T}_z}(s),
\end{equation*}
where specific losses are defined by
\begin{align*}
 \stepManifoldLoss^{\mathcal{T}_x}(s) = &\huber(\relu(\mathcal{T}_{\underline{x}} - \FK_x(\q)) + \\ 
    & \huber(\relu(\FK_x(\q) - \mathcal{T}_{\overline{x}}),\\
    \stepManifoldLoss^{\mathcal{T}_y}(s) = & \huber(\relu(\mathcal{T}_{\underline{y}} - \FK_y(\q)) + \\
    & \huber(\relu(\FK_y(\q) - \mathcal{T}_{\overline{y}}),\\
    \stepManifoldLoss^{\mathcal{T}_z}(s) = & \huber(FK_z(\q) - \mathcal{T}_z),\\
\end{align*}
where $\mathcal{T}_{\underline{x}}, \mathcal{T}_{\overline{x}}, \mathcal{T}_{\underline{y}}, \mathcal{T}_{\overline{y}}$ are the lower and upper boundaries of the table in the $x$ and $y$ directions, while $\mathcal{T}_z$ is the table surface height.

Moreover, we observed that the robot controller has problems with tracking very fast trajectories, especially when trajectory curvature in the workspace is high. Therefore, we introduced an additional regularization term to the typical time-minimization task loss and defined the task loss function by
\begin{equation*}
    \stepNNLoss(s) = 1 + \eta \kappa_{ee}(s) v_{ee}^2(s),
\end{equation*}
where $\eta = 0.01$ is an experimentally chosen regularization factor, while $\kappa_{ee}$ and $v_{ee}$ are respectively the curvature and velocity of the end-effector trajectory.

\subsubsection{Dataset for Air Hockey hitting replanning}
\label{ap:datasets}
The dataset for learning how to hit and be able to replan is similar to the one created just for the hitting task, however, it is far more diversified. To be able to replan, we need to know how to plan between any two configurations in the workspace. Moreover, to be able to perform different trick shots we randomized also the desired hitting direction, initial velocity, and acceleration. The data generation procedure scheme is similar to the one shown in Appendix~\ref{ap:ah_sim_task} 
and differs only in the following steps
\begin{enumerate}[label=\roman*.]
    \item Initial and desired mallet positions range defined by
    $(x,y,z) \in [0.6; 1.3] \times [-0.45; 0.45] \times \{0.16\}$
    \setcounter{enumi}{2}
    \item draw a hitting angle which differs from the direction of the line connecting initial and desired positions no more than $\frac{2\pi}{3}$,
    \setcounter{enumi}{4}
    \item in 80\% of the cases randomly scale the magnitude of the
desired joint velocity, and set to maximal in the rest,
    \setcounter{enumi}{6}
    \item compute random initial joint velocity constrained to the table manifold, or set it to 0 in 20\% of cases,
    \item compute random initial joint acceleration constrained to the table manifold.
\end{enumerate}
The created dataset consists of 120 000 samples, from which 112 000 belong to the training set and the rest to the validation set.

\subsubsection{Parameters of the algorithms used for evaluation}
We compared our proposed approach with several state-of-the-art motion planning algorithms i.e. TrajOpt~\cite{trajOpt}, \gls{mpcmpnet}~\cite{li2021mpcmpnet}, \gls{cbirrt}~\cite{berenson2009manipulation}, and \gls{sst}~\cite{sst}. 
As we aim at increasing the reproducibility of our research, none of these methods was implemented by ourselves, instead, we relied on the following implementations:
\begin{itemize}
    \item for TrajOpt we employed the SLSQP minimization implemented in SciPy, with constraints and cost function implemented by us using pinochchio~\cite{carpentier2019pinocchio} library,
    \item for \gls{mpcmpnet} and \gls{sst} we used the implementation provided by the authors of~\cite{li2021mpcmpnet} with some necessary modifications required to compile and run their code, and our C++ implementation of the heavy object manipulation and robotic Air Hockey systems which also utilizes the Pinocchio library~\cite{carpentier2019pinocchio},
    \item for \gls{cbirrt} we used the OMPL~\cite{sucan2012ompl} implementation and our own constraint and distance definitions.
\end{itemize}
To make our comparison fair we tried to adjust the parameters of these motion planning algorithms, to maximize their performance. Appropriate parameter tuning is especially important for the sampling-based motion planning algorithms, therefore we performed a random search of the parameters to find the regions of the parameter space, which give the highest success ratios and shortest planning times. 
In the case of the \gls{sst} and \gls{mpcmpnet}, some of their parameters are common because they use the same algorithm under the hood, the difference is in the way of selecting the node to expand and in the Expand procedure itself. Therefore, we set for both algorithms the same parameters for node search radius $\delta_{BN} = \SI{0.2}{\m}$, witness radius $\delta_s = \SI{0.1}{\m}$ and goal radius $r_g = \SI{0.2}{\m}$, for the task of heavy object manipulation, and $\delta_{BN} = \SI{0.05}{\m}$, witness radius $\delta_s = \SI{0.02}{\m}$ and goal radius $r_g = \SI{0.02}{\m}$ for robotic Air Hockey hitting. Regarding the "extend" procedure, for both tasks, we set the minimum and the maximum number of steps to 1 and 20 respectively, with an integration step equal $\SI{5}{\ms}$ for \gls{sst}, whereas for \gls{mpcmpnet} we set the parameters of the \gls{cem} MPC solver to 64 samples, 4 elite samples, the maximal number of iterations equal 30 and convergence radius equal 0.02, as they resulted in fast convergence. We have also set the values of the integration step to $\SI{10}{\ms}$, motion time gaussian $(\mu_t, \sigma_t) = (0.05, 0.1)$, with maximal time $t_{max} = \SI{0.1}{\s}$, control gaussian $(\mu_c, \sigma_c) = (\bm{0}, 0.8 \bm{\bar{\tau}})$, for the heavy object manipulation task, for \gls{mpcmpnet}, while for Air Hockey $(\mu_t, \sigma_t) = (0.1, 0.4), t_{max} = \SI{0.5}{\s}, \mu_c, \sigma_c) = (\bm{0}, 0.5 \bm{\bar{\tau}})$. Nevertheless, despite the tuning of the parameters, obtained motion planning times were relatively long, which may be caused by the innate difficulty of the considered problem and very tight constraints, which can be viewed as narrow passages in the configuration space. For both \gls{sst} and \gls{mpcmpnet}, we used the Euclidean distance function in the task space. This simplifies the motion planning problem a lot, as it requires the planners to plan only for the position, disregarding the task of reaching the desired velocity. It is also possible to define a distance function that takes into account the distance in the space of the velocities weighted with the distance in the task space, however, this makes the exploration of the states-space inefficient, slowing down further the planning process.
For \gls{cbirrt}, for both tasks, we set the goal radius to \SI{1}{\cm}, and the allowed tolerance of the constraint to 0.01, while for the range parameter we used an automatically determined value by OMPL.
In the case of the TrajOpt implementation in Scipy, there are no parameters that affect the performance of the method, so we only limited the maximum number of iterations to 100, to avoid prolonged optimization.

Our proposed method also has some important parameters that have to be chosen, however, all of them are parameters of the learning process, as the inference does not depend on anything but data and model. 
For our experiments, we set the number of control points of the $\Paths$ and $\rates$ B-splines to 15 and 20 respectively, as it gives enough freedom to plan accurate trajectories that satisfy constraints. To ensure a high level of trajectory smoothness we set the degree of both B-splines to 7.
Another group of parameters is the one related to the manifold metric optimization i.e. $\bm{\alpha}^{(0)}$. For the heavy object manipulation $\bm{\alpha}^{(0)}$ was set to the zero vector, while for the Air Hockey task we proposed a different initialization, which was meant to equalize the initial gradients of loss functions and was equal to $\bm{\alpha}^{(0)} = \begin{bmatrix} \alpha_{\mathcal{T}}^{(0)} & \alpha_{\dq}^{(0)} & \alpha_{\ddq}^{(0)} & \alpha_{\torque}^{(0)}\end{bmatrix} = \begin{bmatrix} 1 & 1 & 10^{-2}& 10^{-4}\end{bmatrix}$. For the heavy object manipulation task, we defined the following levels of the allowed constraints violations
$\bLM_{E} = 10^{-6}$, $\bLM_{O} = 10^{-5}$, $\bLM_{\dq} = 6\cdot10^{-3}$, $\bLM_{\ddq} = \bLM_{\torque} = 6\cdot10^{-2}$, while for the hittng task $\bLM_{\mathcal{T}} = 2\cdot10^{-6}$, $\bLM_{\dq} = 6\cdot10^{-3}$, $\bLM_{\ddq} = 6\cdot10^{-2}$, $\bLM_{\torque} = 6\cdot10^{-1}$. The update step for the $\alpha$ parameters was set to $10^{-2}$ for both motion planning problems. The learning step for neural network parameters was set to $5\cdot10^{-5}$, while the batch size was set to $128$.

\addtolength{\textheight}{-10cm} 

\begin{IEEEbiography}[{\includegraphics[width=1in,height=1.25in,clip,keepaspectratio]{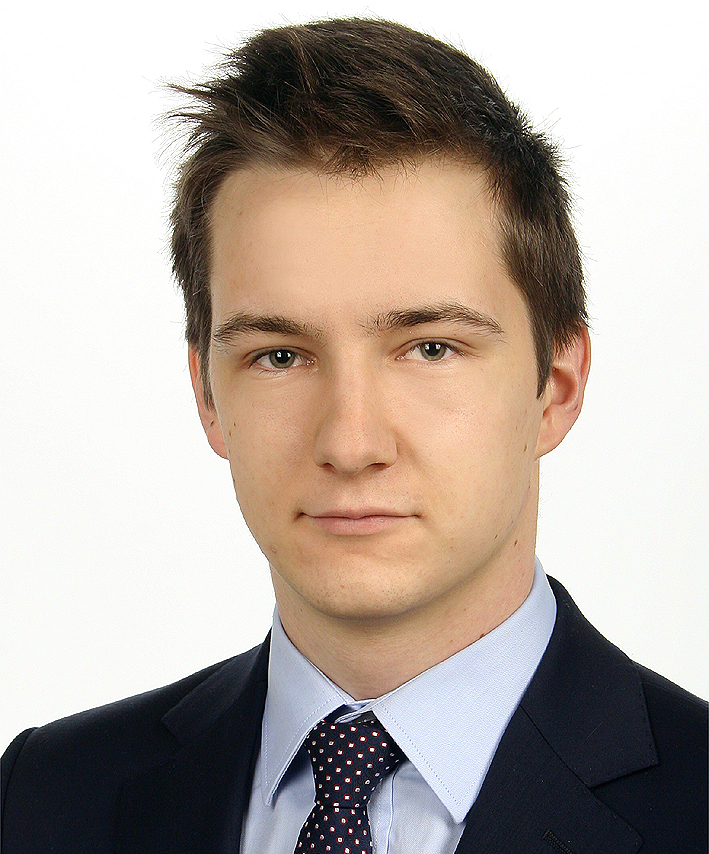}}]{Piotr Kicki} received his B.Eng. and M.Sc. degrees in automatic control and robotics from Poznan University of Technology, Poland in 2018 and 2019, respectively. He is currently a Ph.D. student in Robotics, at the Institute of Robotics and Machine Intelligence at Poznan University of Technology, Poland. His main research interests include robot motion planning and applications of machine learning in robotics.
\end{IEEEbiography}

\begin{IEEEbiography}[{\includegraphics[width=1in,height=1.25in,clip,keepaspectratio]{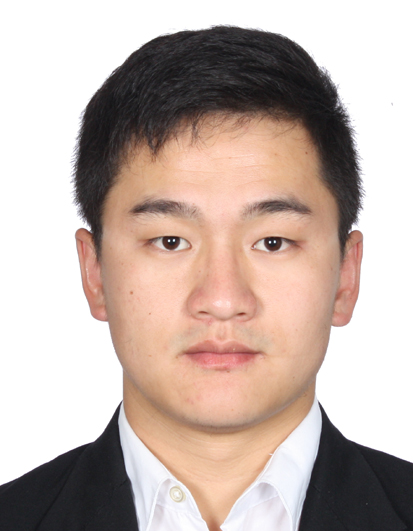}}]{Puze Liu}
is pursuing his Ph. D. degree at Intelligent Autonomous Systems Group, Technical University Darmstadt since 2019. Prior to this, Puze received his M. Sc. in Computational Engineering from Technical University Berlin and B. Sc from Tongji University, China. Puze's research interest lies in the interdisciplinary field of robot learning that tries to integrate machine learning techniques into robotics. His prior work focuses on optimization, control, reinforcement learning, and safety in robotics.
\end{IEEEbiography}

\begin{IEEEbiography}[{\includegraphics[width=1in,height=1.25in,clip,keepaspectratio]{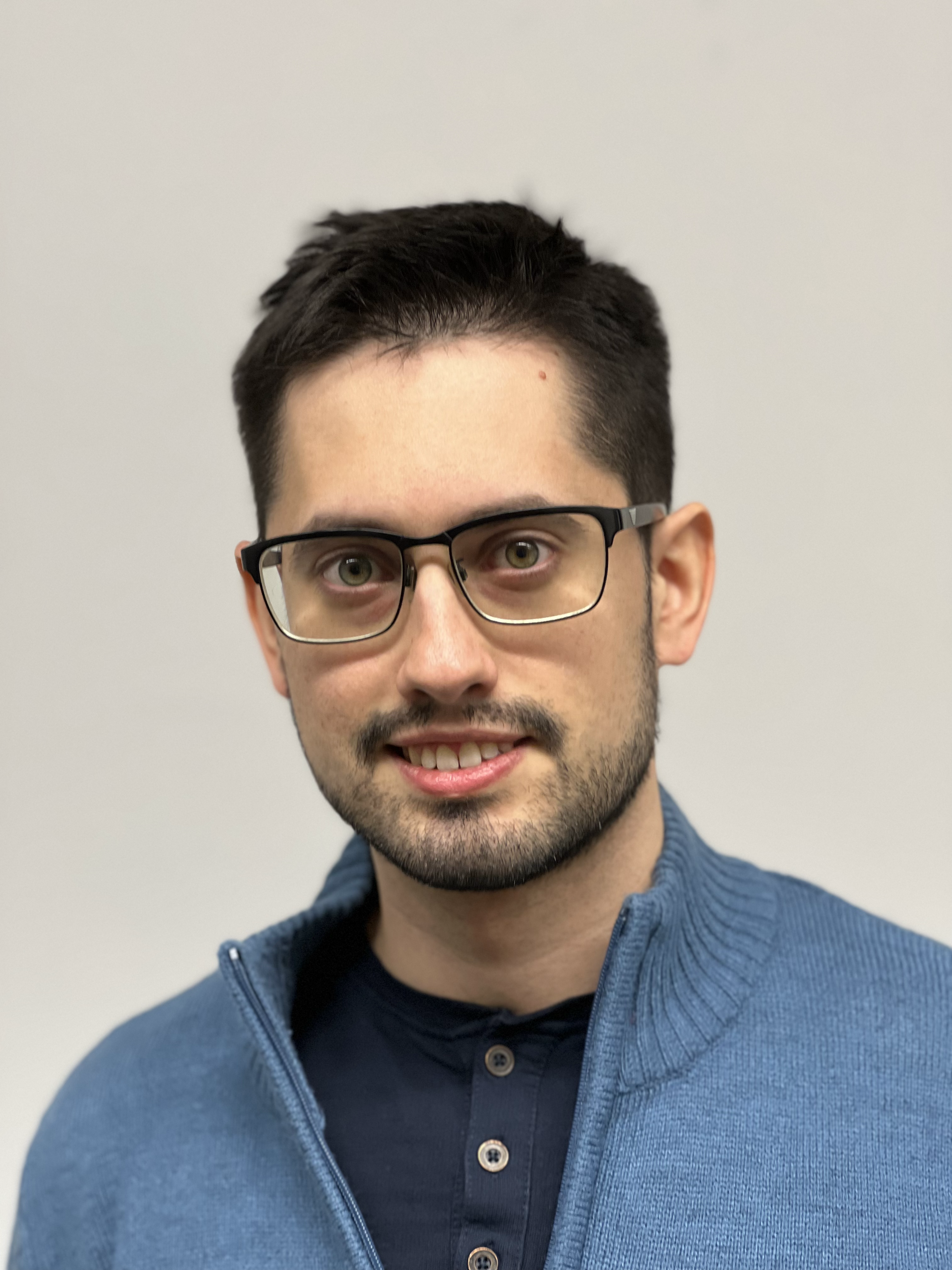}}]{Davide Tateo}
is a Research Group Leader at the Intelligent Autonomous Systems Laboratory in the Computer Science Department of the Technical University of Darmstadt. 
He received his M.Sc. degree in Computer Engineering at Politecnico di Milano in 2014 and his Ph.D. in Information Technology from the same university in 2019. 
Davide Tateo worked in many areas of Robotics and Reinforcement Learning, including Deep Reinforcement Learning, Planning, and Perception.
His main research interest is Robot Learning, focusing on high-speed motions, safety, and interpretability.
He is one of the co-authors of the MushroomRL Reinforcement Learning library.
\end{IEEEbiography}

\begin{IEEEbiography}[{\includegraphics[width=1in,height=1.25in,clip,keepaspectratio]{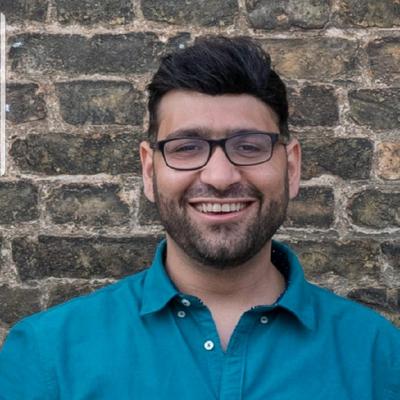}}]{Haitham Bou-Ammar}
leads the reinforcement learning team at Huawei technologies Research \& Development UK and is an Honorary Lecturer at UCL. 
His primary research interests lie in the field of statistical machine learning and artificial intelligence, focusing on Bayesian optimization, probabilistic modeling, and reinforcement learning. He is also interested in learning using massive amounts of data over extended time horizons – a property common to "Big-Data" problems. His research also spans different areas of control theory and nonlinear dynamical systems, as well as social networks and distributed optimization.
\end{IEEEbiography}

\begin{IEEEbiography}[{\includegraphics[width=1in,height=1.25in,clip,keepaspectratio]{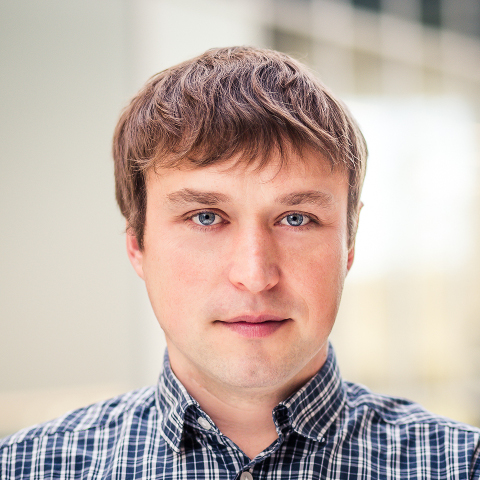}}]{Krzysztof Walas} graduated from Poznan University of
Technology (PUT) in Poland receiving M.Sc. in Automatic Control and Robotics. He received (with honours)
Ph.D. in Robotics in 2012 for his thesis concerning
legged robots locomotion in structured environments.
Currently, he is an Assistant Professor at the Institute
of Robotics and Machine Intelligence at PUT, Poland.
His research interests are related to robotic perception
for physical interaction applied both to walking and
grasping tasks.

\end{IEEEbiography}

\begin{IEEEbiography}[{\includegraphics[width=1in,height=1.25in,clip,keepaspectratio]{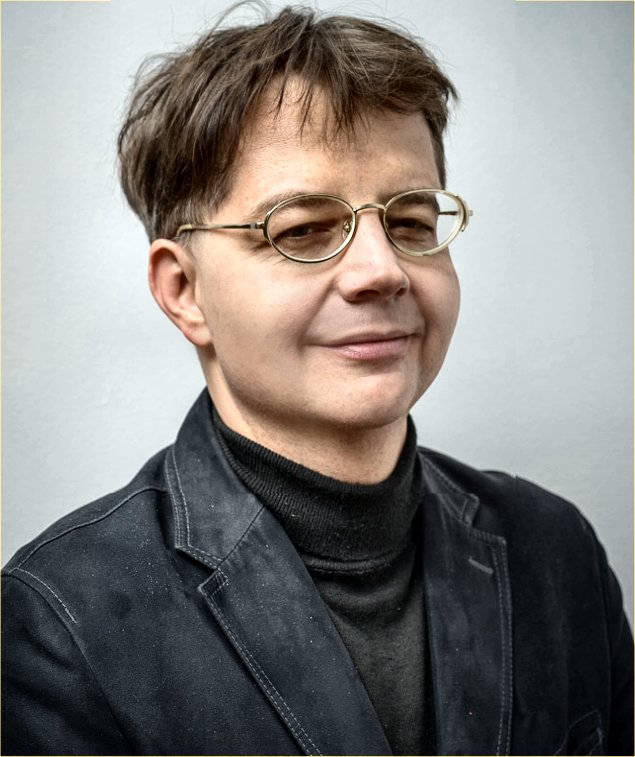}}]{Piotr Skrzypczyński}
is a full professor in the Institute of  Robotics and Machine Intelligence (IRIM) at Poznan University of Technology (PUT) and head of the IRIM Robotics Division. Piotr Skrzypczy\'nski received the Ph.D. and D.Sc. degrees in Robotics from PUT in 1997 and 2007, respectively. He authored or co-authored more than 160 technical papers in robotics and computer science. His current research interests include AI-based robotics, robot navigation and SLAM, computer vision, and machine learning.
\end{IEEEbiography}

\begin{IEEEbiography}[{\includegraphics[width=1in,height=1.25in,clip,keepaspectratio]{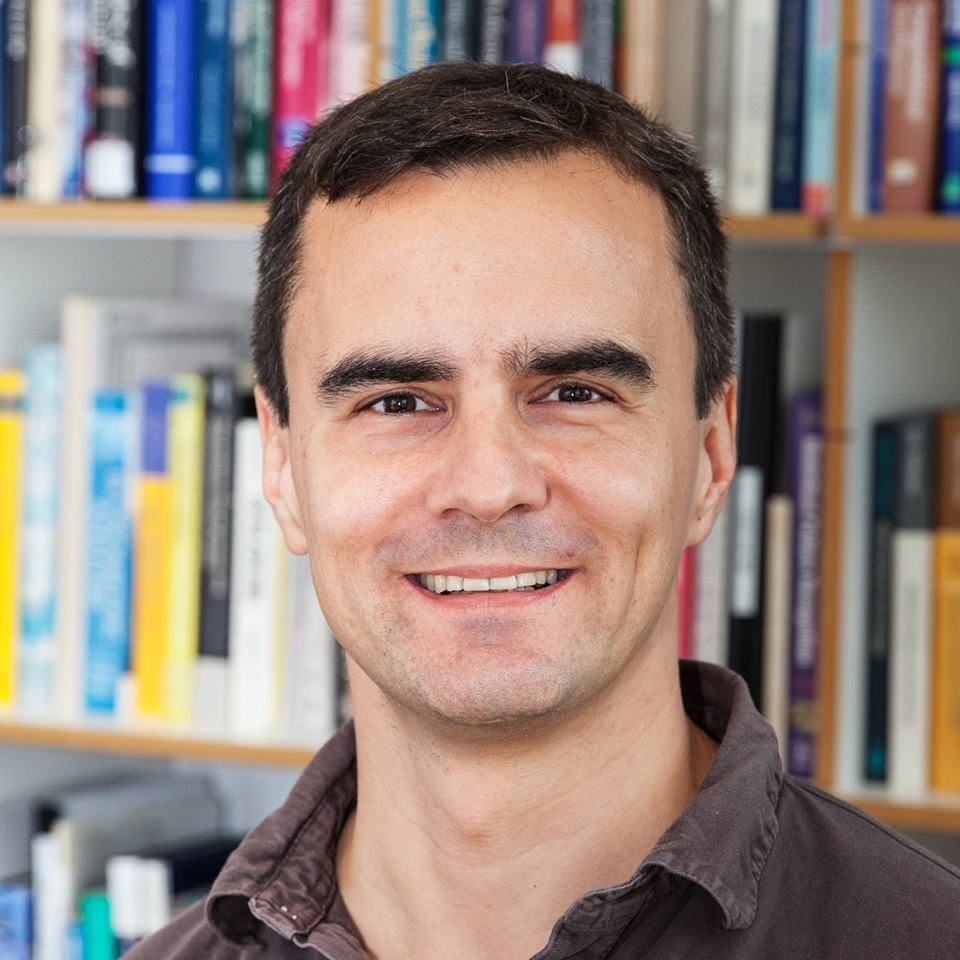}}]{Jan Peters}
is a full professor (W3) for Intelligent Autonomous Systems at the
Computer Science Department of the Technische Universitaet Darmstadt.
Jan Peters has received the Dick Volz Best 2007 US Ph.D. Thesis Runner-Up Award, the Robotics: Science \& Systems - Early Career Spotlight, the INNS Young Investigator Award, and the IEEE Robotics \& Automation Society's Early Career Award as well as numerous best paper awards. In 2015, he received an ERC Starting Grant and in 2019, he was appointed as an IEEE Fellow.
\end{IEEEbiography}

\vfill

\end{document}